# Xapagy: a cognitive architecture for narrative reasoning

Ladislau Bölöni

Dept. of Electrical Engineering and Computer Science

University of Central Florida

Orlando, FL 32816–2450

lboloni@eecs.ucf.edu

*Abstract*—We introduce the Xapagy cognitive architecture: a software system designed to perform *narrative reasoning*. The architecture has been designed from scratch to model and mimic the activities performed by humans when witnessing, reading, recalling, narrating and talking about stories.

## Contents

## I. INTRODUCTION

*It was not only difficult for him to understand that the generic term **dog** embraced so many unlike specimens of differing sizes and different forms; he was disturbed by the fact that a dog at three-fourteen (seen in profile) should have the same name as the dog at three-fifteen (seen from the front).*

("Funes the memorious" by Jorge Luis Borges)

This paper describes the Xapagy cognitive architecture. The system is designed to perform *narrative reasoning*, an activity roughly analogous to some of the mental processes humans perform with respect to stories. Narrative reasoning includes:

- Witnessing a series of ongoing events (a story), keeping track of the participants, their identity, properties and activities.
- Following a fixed story narrated in a language (for instance, by reading a text or listening to another agent's narration).
- Predicting future events in the story, expressing surprise when unexpected events occur.
- Inferring events which (for some reason) were not witnessed; understanding narrations where some events have been not explicitly said ("reading between the lines").
- Recalling a story, summarizing or elaborating on the remembered story, chaining together remembrances.
- Daydreaming, confabulating new stories.
- Self-narrate the story by verbalizing the recalled or confabulated story, for the narrating agent's own use.
- Narrate the story to an audience, adapt the narration based on feedback from an audience, elaborate on aspects of the story or selectively narrate.
- Act as an audience for a narration, express surprise or puzzlement, request clarification or elaboration for parts of the story and ask questions.
- Perform collaborative story-telling, develop a story by alternating narrations from multiple agents.

The Xapagy architecture has been developed from scratch, rather than as an evolution of an existing model from artificial intelligence or cognitive science (but, naturally, building on the experience of these systems). Starting from scratch was motivated partially by the fact that the targeted behavior only partially overlaps with that of current cognitive architectures. The other reason, however, was an attempt to push the limits of the computational performance. We require that the reasoning step (such as the reading, witnessing, recalling or generating a sentence) to be performed in *constant time* on a computer with an appropriate level of parallelism. The cumulative computational complexity of all the algorithms used to perform the reasoning step must be $O(1)$.

This is an unusually restrictive requirement for any algorithm in computer science, and artificial intelligence in particular. For instance, in computational logic, an algorithm operating in $O(n^3)$ might be considered fast, and intractable problems are legitimate subjects of inquiry. It is true, using anytime algorithms [30] can turn any architecture into a constant time architecture. We found it useful, however, to search for algorithms which are naturally of low complexity, rather than using high complexity algorithms with anytime hooks. In this endeavor we are encouraged by the fact that the human brain can perform narrative reasoning with only about a dozen synapses separating input from output [27].

Describing a cognitive architecture built from scratch inevitably requires a long list of definitions of primitive components. We anticipate that the most challenging aspects for the reader will be "false friends": terms such as concept, instance, identity, verb and focus, which in the Xapagy system have somewhat unexpected definitions. To allow the reader to grasp the overall picture before proceeding to a detailed description, Section II describes the organization of the Xapagy architecture in a top down, informal way. This section will also include a short tutorial to the Xapagy pidgin language (Xapi). Section III describes the primitive components of the Xapagy system and their elementary operations. Section IV focuses on *story following* and *shadowing* which is the basic reasoning mechanism in Xapagy. Section V describes the representation of problem domain knowledge in Xapagy, as well as the core knowledge shared by all Xapagy agents. This includes the representation of attributes, change, ownership, grouping and scenes. Section VI discusses the representation of episodic knowledge. Section VII discusses the role of the Xapi language in the narrative process and issues of reference resolution, reference selection and verbalization. Section VIII discusses the implementation of narrative reasoning such as recalling, confabulating and narrating stories, as well as agent interaction with regards to story telling. Section IX positions the Xapagy system relative to other cognitive architectures and intelligent systems. Section X concludes the paper.

Appendix A describes the notations and typographical conventions used in the paper.

**Note:** this paper describes the *architecture* of the Xapagy software system. Due to space limitations, we postponed for future publications the full detail description of such components such as the shadow and headless shadow maintenance (Section IV-C and Section VIII-A) and the detailed implementation of different recall and confabulation models (Section VIII-C). Some of the examples in the paper have been hand-engineered for illustration purposes - usually by simplifying the real output and internal state of the system. However, all the components described in the paper are implemented and (at least to some degree) functional. Functionality which is planned but not implemented is noted as such in the paper. Whenever the paper refers to the "current version of the Xapagy system", it means core system Xapagy v0.9.32 with the language model Xapi v3.2 and domain knowledge library XapagyDKL v0.20.

## II. A TOP-DOWN INTRODUCTION

### A. External look: the pidgin language

The Xapagy architecture describes the operation of an autonomous agent which can directly witness events happening in the world, and it can communicate with humans and other agents through the *Xapi pidgin language*.

In linguistics, *pidgin languages* [26] are natural languages with a simplified syntactic structure which appear when two groups of people need to communicate without the time necessary to properly learn each other's languages. Pidgin languages are not the native language of any group of people, and uniquely among human languages, they have a limited expressiveness. Pidgins normally evolve into full featured *creole* languages when they are used as first language by a population. For instance, although it originated as a pidgin, Tok Pisin, official language of New Guinea, is not a pidgin in linguistic terms, but a creole, a language capable of expressing the full range of human concerns. Despite their limitations, pidgin languages represent a useful stopgap measure for communication between human communities, and their creative use can in fact express a wide range of information.

We have chosen to model the language of our agents on pidgin languages for their desirable properties both on the human and the Xapagy agent side.

- **The human side:** Xapi should be immediately understandable to English speakers. Some simple English sentences can be

immediately translated to Xapi. For other sentences, the missing grammatical structures can be approximated with workarounds. Although we make no claim that the full meaning of English text can be translated to Xapi, many stories as well as interaction patterns can be captured in a satisfactory manner.

- **The computer side:** The simplified syntax of Xapi allows us to bypass the complexities of natural language processing, while still allowing communication with humans.

Xapi shares some important features with human pidgin languages. It has an uncomplicated causal structure: the only supported compound statement is the quotation statement. It uses separate words to indicate degrees of properties. It does not support quantifiers. It has a fixed word order and no morphophonemic variation (more exactly, accepts a range of morphophonemic variants as synonyms from the human speaker, but it does not provide them when generating it from the computer side[1]). In addition, it uses sentence part separators ("/") and sentence boundary markers ("//") as a way of circumventing the necessity of complex text segmentation (which is beyond the objectives of the Xapagy system).

Xapi is intended to be read and written by humans when communicating with Xapagy agents. It is also used for communication between Xapagy agents, but it is definitely *not* a formal agent communication language.

A line of Xapi text represents a single sentence, with the sentence parts separated by "/" and terminated with a period "." or question mark "?". Sentences can be of a subject-verb-object form:

```
1  The boy / hits / the dog.
```

subject-verb form:

```
1  The boy / cries.
```

subject-verb-adjective form:

```
1  "Hector" / is-a / warrior.
```

or verb instance-verb-adverb form:

```
1  "Achilles" / strikes / "Hector".
2  The strikes / action-is / hard.
```

One or more parts can be substituted with a `wh` component, transforming the sentence into a question:

```
1  Wh / eats / "Little Red Riding Hood"?
```

Xapi supports a single form of compound sentence, the *quotation sentence*:

```
1  "Little-Red-Riding-Hood" / says / conversation //
2    the eyes / is-a / big.
```

In some cases, the semantics of other compound or complex sentences can be approximated by sentences which refer to shared instances or verb instances. We make, however, no claim that the expressive power of Xapi matches that of a natural language.

Subjects and objects are *instances* which are either currently in the *focus*, or are newly created by the sentence. A new instance can be created by prefixing a word with the indefinite article "a/an":

```
1  "Billy" / hits / a dog.
```

In this example we assume Billy has been referred to before, but the dog has been just introduced in the story. Subsequent references to the already introduced instance of the dog are prefixed with the definite article "the" (which can be omitted for proper nouns).

```
1  The dog / changes / angry.
2  The dog / bites / "Billy".
```

In pidgin, we refer to instances through one or more of their *attributes*. When we mention the attribute `[dog]`, the reference will be made to the strongest instance in the scene which has the given attribute. In some cases, such as quotation sentences, the resolution process is performed in a different scene. It is the responsibility of the speaker to choose attributes which make a reference resolve correctly.

[1]In the current version Xapi 3.2.

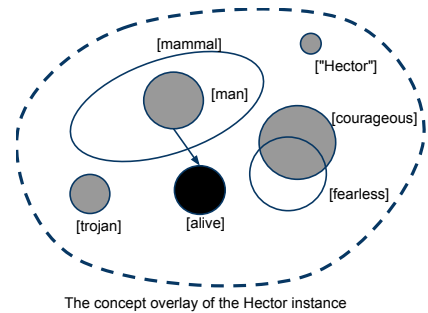

Fig. 1. A visualization of concepts, overlap, impact and concept overlays in the form of patches. Concepts directly added are represented with gray patches, concepts added through impact are black patches, and concepts which are implicitly present in the overlay due to their overlap with explicitly added concepts are represented as transparent contours.

The verb word in a Xapi sentence actually maps to a mixture (overlay) of *verb concepts* in the internal representation of the Xapagy agent. The composition of this verb overlay determines the relationship between the sentences. For actions such as "hits" or "bites", the relationship between the sentences is one of a *weak temporal succession*. One can imagine these sentences connected by "then": Billy hits the dog, then the dog is angry, then the dog bites Billy. The relationship is stronger between sentences which share instances.

We can create verb overlays which convey essentially the same action but create a stronger succession to the preceding verbs by adding the word "thus". This can be used to represent a cause-effect relationship:

```
1  "Billy" / hits hard / a dog.
2  The dog / thus changes / angry.
```

Just like the "dog" in this story is an instance which has as attributes *concepts* such as `[dog]` and `[angry]`, the action "hit" is a verb instance whose verb part has as attributes the *verb concepts* `[hit]` and `[hard]`.

Finally, there are some sentences which do not represent actions in time, and thus they are not connected by succession relationships. Examples are verbs which set attributes to instances:

```
1  "Hector" / is-a / warrior.
```

or establish relationships between instances:

```
1  "Hector" / loves / "Andromache".
```

*B. From words to concepts and verbs*

We have seen that the Xapi pidgin uses a simplified syntax, but otherwise regular English words. The *dictionary* of the agent maps nouns and adjectives to *overlays of concepts* while verbs and adverbs are mapped to *overlays of verb concepts*. We will discuss concept overlays, as the verb overlays are very similar.

An overlay is the simultaneous activation of several concepts with specific levels of energy. For instance the dictionary of a Xapagy agent might associate the word "warrior" with the following overlay: `[courageous=0.4, violent=0.3, strong=0.3]`

The attributes of an instance are represented by an overlay which can be gradually extended through the side effects of the sentences. Thus, when reading the Xapi sentences:

```
1  "Hector" / is-a / man.
2  "Hector" / is-a / warrior.
```

the instance identified with the attribute Hector will acquire the attributes described in the overlay: man, courageous and so on.

Concepts are *internal* structures of the Xapagy agent. To distinguish them from *words*, which are external entities, we will always show them in brackets, such as `[warrior]`.

One way to develop an intuition for concepts and overlays is to visualize them as patches of a certain area in a fictional two dimensional space. Some concepts have a large area (e.g. `[animal]`), while others are smaller (e.g. `[dog]`). Proper nouns such as `["Hector"]` have a very small area. Overlays can be visualized as collections of such patches (see Figure 1).

Concepts can *overlap* on a pair-by-pair basis. For instance, there is a full overlap between man and human, meaning all men are human: *overlap*(`[man]`,`[human]`) = *area*(`[man]`). Thus, if we inquire whether Hector is human, we shall obtain a value of 1.0. There is, on the other hand, only a partial overlap between courageous and fearless: *overlap*(`[fearless]`, `[courageous]`) = $0.5 \cdot area$(`[courageous]`). Thus, is we ask whether Hector is fearless, the answer will be $0.4 \times 0.5 = 0.2$.

Words denoting proper nouns, such as "Hector", marked in pidgin by quotation marks, are treated slightly differently: when the agent first encounters a proper noun, it will create a new concept with a very small area, and an entry in the domain dictionary associating the proper noun with an overlay containing exclusively the new concept. Other than this, proper nouns are just like any other attributes. Having the same proper noun as an attribute does not immediately imply any form of identity.

The dictionary which maps from a word to an overlay, the areas and overlap of the concepts are part of the *domain knowledge* of the agent. Different agents might have different domain knowledge - thus the meaning of the word might differ between agents.

### C. Instances

The definition of an instance in Xapagy is somewhat different from the way this term is used in other intelligent systems. Instead of representing an entity of the real world, it represents an *entity of the story, over a time span limited by the additivity of the attributes*. For a particular instance, its attributes, represented in a form of an overlay of concepts, are additive: once an instance acquired an attribute, the attribute remains attached to the instance forever.

The advantage of this definition is that once we have identified an instance, there is no need for further qualification in order to identify its attributes (nor its actions).

What might be counter-intuitive for the reader, however, is that things we colloquially call a single entity are represented in Xapagy by several instances. Let us, for instance, consider Hector. In the Iliad, he is a central figure of the story: he appears first as a warrior, participates in several fights, while later he is killed by Achilles by a spear to his throat, and the action revolves around the return of his body to his father, Priam. Hector also appears in the Hollywood movie "Troy", but here he is killed with a sword to the chest. In the science fiction novel "Ilium" by Dan Simmons, Hector is quantum-recreated by aliens to replay the events in Iliad on the planet Mars. In the novel, Hector chooses to ally itself with Achilles and the Greeks against the aliens from an alternate reality who are playing the roles of the gods.

In the Xapagy system, these are all different instances, which share the attribute `["Hector"]` (but then, that is also shared by Hector Berlioz, the French composer). These instances, of course, are connected through various relations of identity (for a discussion on the philosophical problem of personal identity we refer the reader to [20], [21]). Such identity relations are represented in Xapagy by *relations* among multiple instances, *not* by sharing the same instance.

The Xapagy system, however, moves a step beyond this. Not only Hector from the Iliad and Hector from the Ilium are represented by different instances, but Hector the live warrior and Hector the corpse in the Iliad are also two different instances, as the change from a living warrior to a dead one can not be represented as an addition of attributes.

```
1   "Achilles" / strikes / "Hector".
2   "Hector" / thus changes / dead.
3   "Achilles" / kicks / "Hector".
```

The sentence in line 2 will create a new instance, with a new set of attributes. In addition, it will connect these instances with a somatic identity relation:

```
1   i201 ["Hector"] / is-somatically-identical /
        i101 ["Hector"].
```

Whenever, at a later time, the Xapagy agent recalls Hector (for instance, in a conversation) it first needs to establish what instance is under consideration. Once this instance has been unequivocally established to be, for instance, the live Hector `i101`, all the attributes of the instance are also unambiguously established: we can say that he is strong, courageous etc., attributes which would not make sense applied to the dead Hector instance `i201`.

### D. The focus

The focus in the Xapagy system holds instances and verb instances after their creation for a limited time interval during which they are *changeable*. After an instance or verb instance leaves the focus, it can never return - and thus, it remains unchanged.

Instances in the focus can acquire new attributes, participate as a subject or object in verb instances and become part of relations. Verb instances in the focus can become part of succession or summarization relations, and they can be referred to by new verb instances.

A visual thinking oriented reader might think about the focus in the following way: the focus is a dynamically evolving graph. New nodes (instances and verb instances) are added through various events. The same events might also create new edges among the nodes of the focus. When a node leaves the focus, it retains its attributes and edges, but it can not acquire new ones any more. So the focus can be seen as the actively growing boundary of a graph which represents the complete experience of the Xapagy agent. The graph will be only *locally connected*: it will *not* have long links, as only nodes which have been in the focus together can have links.

To illustrate this graph nature of the focus, let us consider the following short story:

```
1   A "Billy" / is-a / boy.
2   A "Jeannie"/ is-a / girl.
3   "Billy" / kissed / the girl.
4   "Jeannie" / pushed / the boy.
5   The boy / laughed.
```

Running this story through a Xapagy agent, and visualizing the focus at the end of the story through the Xapagy system's GraphViz based output, we obtain Figure 2. Such a graphical output provides an insight in the complexity of the focus, but it makes for a difficult reading. Thus, in the reminder of the paper, we shall mostly rely on text based illustrations of the internal structures (see Appendix A for some of the typographical conventions we shall use).

### E. Shadows

Instances and verb instances leaving the focus will be *demoted to the memory* of the Xapagy agent with a certain level of *salience*. They will never enter the focus again. On the other hand, each instance and verb instance in the focus has a *shadow*, a weighted collection of instances and, respectively, verb instances from the memory.

The shadows are maintained through a combination of techniques whose goal is to make the shadows consistent between each other and match the ongoing story with the collections of stories in the shadows. To ensure the matching of the stories, the shadows sometimes need to forego the individual matching level between the instances. Let us consider that a Xapagy agent which had read about the duel between Hector and Patrocles:

```
1   "Hector" / cuts / "Patrocles".
2   The greek / thus changes / dead.
```

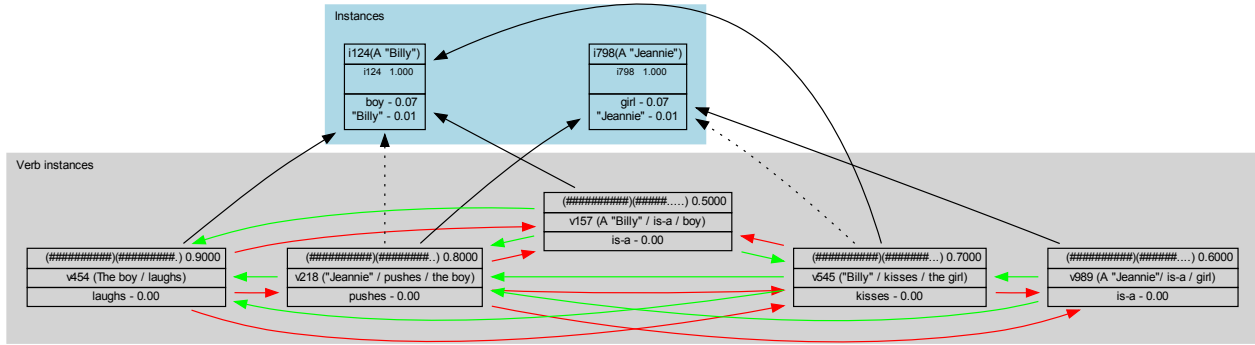

Fig. 2.   Example of the focus after the boy-kisses-girl story.

Several days later, he resumes reading the Iliad and reads:

```
1   "Achilles" / strikes / "Hector".
```

We already know that the two instances of Hector will not be the same: the days passed before resuming the story will be more than enough to remove the instance from the focus. Yet, the two instances will have a lot of common attributes: Hector, Trojan, warrior.

Yet the overall shape of the current fight will lead to a different shadowing: the strongest instance in the shadow of Hector will be the previous instance of Patrocles, while the shadow of Achilles will contain the previous instance of Hector. The Achilles-strikes-Hector verb instance will be shadowed by the Hector-cuts-Patrocles verb instance.

Shadows are always matched to instances and verb instances in the focus. The verb instances in shadows, however, bring with themselves verb instances to which they are connected through succession, precedence and summarization relations. These verb instances can be clustered into weighted sets which are very similar to shadows, but they are *not* connected to current focus components. These sets are called *headless shadows* and they represent outlines of events which the agent either expect to happen in the future or assume that they had already happened but have not been witnessed (or they are missing from the narration). If an event matching the headless shadow happens, the two are combined to become a regular focus-component / shadow pair.

Shadowing is the fundamental reasoning mechanism of the Xapagy architecture. All the higher level narrative reasoning methods rely on the maintenance of the shadows. For instance, the Xapagy agent can predict the death of Hector through the shadow, and can express surprise if this does not happen. While in this example the shadow is created after the events are inserted into the focus from an external source (for instance, by reading), the opposite is also possible. In the case of recall, narration, or confabulation, the agent creates instances and verb instances in the focus based on pre-existing shadows.

With this, we conclude our informal introduction of the Xapagy architecture. The remainder of the paper will present a more detailed description of how these components are implemented.

## III.   THE PRIMITIVE COMPONENTS OF THE ARCHITECTURE

### A.   Random identifiers

The Xapagy system operates by adding verb instances (continuously) and instances (occasionally) to the focus. The need for new instances is fulfilled by a generator which supplies a continuous flow of random identifiers. When a new instance or verb instance is needed, the Xapagy agent picks the current identifier from the continuous flow, and introduces it into the focus.

Identifiers which will become instances are *decorated* with attributes in the form of a concept overlay, and can participate in relationships appropriate for instances (such as ownership, group membership and scene participation). Identifiers which become verb instances are decorated with attributes in form of a verb concept overlay, can participate in sentence part relationships (subject, object and so on), and can be parts of relationships between verb instances (succession and summarization).

The identifiers *do not carry meaning*. However, for the purpose of debugging or illustration we can force the random generator to create descriptive identifiers. In this paper, for easier readability, identifiers such as `i001`, `i002`... will represent instances while `v001`, `v002` ... will represent verb instances. When describing several consecutive stories, we shall jump with 100 in the count of identifiers for every new story.

### B.   Concepts and overlays

A *concept* in Xapagy is the representation of an *undivisible attribute*. A weighted superposition of concepts is called a *concept overlay* (CO). When talking about concepts in general, we will denote concepts with $c_1, c_2 \ldots$ and COs as $C_1, C_2 \ldots$. For specific concepts we will use descriptive names in brackets such as `[man]`. For specific overlays, we list the participating concepts inside brackets, if necessary, specifying the explicit energy level of each concept in the overlay.

The *specificity* of a concept is characterized by its *area*: $area(c) \in \mathbb{R}^+$. The more specific a concept is, the smaller its area. We will assign an area of 1.0 to the concepts corresponding to the *basic objects* of the hierarchy in the sense described in Rosch *et al.* [23]. For instance, some areas used in our experiments are:

```
area([wolf]) = 1.0
area(["Hector"]) = 0.1
area([animal]) = 3.0
area([thing]) = 10.0
```

In the following, we introduce formulas for the calculation of the energy level of specific concepts in overlays. To simplify the formulas we will define a *trimming* function as follows:

$$trim(x, y) = \begin{cases} 0 & \text{if } x < 0 \\ x & \text{if } x \in [0, y] \\ y & \text{if } x > y \end{cases} \quad (1)$$

where we will omit $y$ when its value is 1.

Overlays are built iteratively, by adding one concept at a time, starting with an empty overlay. We start by defining the *explicit energy* of concept $c$ in overlay $C$ as $een(C, c) < area(c)$. In the empty overlay, the explicit energy of all concepts is zero, for non-empty overlays, the explicit energy is determined recursively by the formulas defining the addition operation.

We will define the explicit energy of the overlay as the sum of the explicit energies of all the concepts in the overlay:

$$een(C) = \sum_c een(C, c) \quad (2)$$

We define two addition operations, differentiated by whether they consider or not the impact of the concepts.

The *direct addition* $\oplus$ is a simple summation of the explicit energy, limited by the area of the concepts, and is defined through the following formulas:

$$C' = C \oplus \{c, e\} \Rightarrow$$
$$een(C', c) = trim(een(C, c) + e, area(c))$$
$$\forall c_x \neq c \;\; een(C', c_x) = een(C, c_x) \quad (3)$$

The second, *impacted addition* operation $\boxplus$ also considers the *impact* between concepts. Adding a concept $c_1$ with energy $e$ through this operation automatically triggers the addition of a number of other concepts, with an energy proportional with $e$, defined by the value $impact(c_1, c_2)$ which has the dimensionality of a positive or negative ratio. The impact is not necessarily symmetrical and it can be negative.

The impacted addition operation $\boxplus$ is defined through the formulas:

$$C' = C \boxplus \{c, e\} \Rightarrow$$
$$een(C', c) = trim(een(C, c) + e, area(c))$$
$$\forall c_x \neq c \;\; een(C', c_x) =$$
$$trim(een(C, c_x) + e \cdot impact(c, c_x), area(c_x)) \quad (4)$$

The explicit energy of a concept in an overlay is the energy we explicitly added either through direct addition or through impact.

Another way through which a concept can have energy in an overlay is through *overlapping* a concept which has explicit energy. The overlap between two concepts $c_i$ and $c_j$ is defined in the dimensionality of the area $overlap(c_i, c_j) \in \mathbb{R}^+$. The overlap is always smaller than the area or either concept $area(c_i) \geq overlap(c_i, c_j)$ and it is symmetrical $\forall i \forall j \; overlap(c_i, c_j) = overlap(c_j, c_i)$. If two concept's areas are identical and their overlap is identical with the area, the two concepts are indistinguishable, and thus are considered equivalent:

$$area(c_1) = area(c_2) = overlap(c_1, c_2) \overset{def}{\Longrightarrow} c_1 \equiv c_2 \quad (5)$$

We define the *energy* of concept $c$ in overlay $C$ conservatively by:

$$en(C, c) =$$
$$\max\Big(area(c), een(C, c) +$$
$$\max_{c_x \neq c}\Big(overlap(c_x, c) \cdot \frac{een(C, c_x)}{area(c_x)}\Big)\Big) \quad (6)$$

The advantage of this conservative metric (i.e. relying on a max rather than a sum) is that the areas shared by three or more concepts do not need to be considered, as they do not enter into the calculation of the energy levels of the individual concepts in an overlay.

For a certain overlay $C$ and concept $c$ we can define the *activation* $act(C, c) \in [0, 1]$ of the concept in the overlay with:

$$act(C, c) = \frac{en(C, c)}{area(c)} \quad (7)$$

Having defined the formulas governing concepts and COs, let us investigate the intuitions behind them. The definition of a concept and related terms in Xapagy does *not* map to scientific classification, computational logic, descriptive logic, or the possible worlds interpretation. The Xapagy system, in its current version, can not model these abstractions[2].

Roughly, Xapagy concepts cover the categories of nouns *and* adjectives of classical grammar. There is no notion of a "class" or "type" in Xapagy: there are only instances which happen to share certain attributes. A man is simply a random identifier which happened to have the attribute `[man]`.

The area of a concept, as we said is a metric of its specificity - more general concepts have a larger area. However, we should caution the reader against pushing the analogy between this metric and the area of two-dimensional shapes too far. For instance, we can not assume that the sum of the areas of the different animal types will be the area of the concept `[animal]`.

In a similar vein, Xapagy verbs cover both the categories of verbs *and* adverbs of classical grammar. A certain verb word in a sentence maps to an overlay of attributes. An interesting future research direction could be to investigate the relationship between this composition model and Pinker's *microfeatures* [22].

### C. Negation

For each concept $c$ we automatically define its negation $-c$. When we write out the name of the concept, such as in `[alive]`, we shall write the negation as `[not-alive]`. The negated concept is defined by its specific impact and overlap with reference to other concepts.

$$impact(c_1, -c_1) = impact(-c_1, c_1) = -1$$
$$overlap(c_1, c_2) = overlap(-c_1, -c_2)$$
$$impact(c_1, c_2) = impact(-c_1, -c_2) \quad (8)$$

The negation defined in Xapagy does not follow logical (or arithmetical) rules of negation. The concept `[alive]` does not stand for all the instances which are alive, nor `[not-alive]` stay for all the other instances.

The definition of the negation, however, is designed to serve the needs of narrative reasoning, and most of the times yields results consistent with the commonsense interpretation:

```
1   "Hector" / is-a / warrior.
2   He / is-a / alive.
3   "Hector" / changes / not-alive.
```

The new instance of Hector will have the attributes `[Hector, not-alive]`, the `[alive]` attribute being removed by the negative impact of the `[not-alive]` concept[3].

---

[2]If such abstractions must become part of the narrative, they need to be modeled in the story itself.

[3]We can, of course, define a Xapi word "dead" which maps to `[not-alive]`.

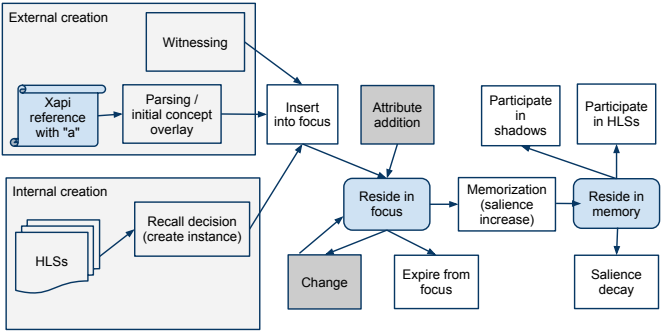

Fig. 3.    The life cycle of an instance.

## TABLE I
### The verb instance types and their mandatory parts

| Type | Verb | SubI | ObjI | SubVI | ObjCO | ObjVO |
|------|------|------|------|-------|-------|-------|
| S-V-O | x | x | x | - | - | - |
| S-V | x | x | - | - | - | - |
| S-ADJ | x | - | - | - | x | - |
| S-ADV | x | - | - | - | - | x |
| IND | x | x | x | - | - | x |

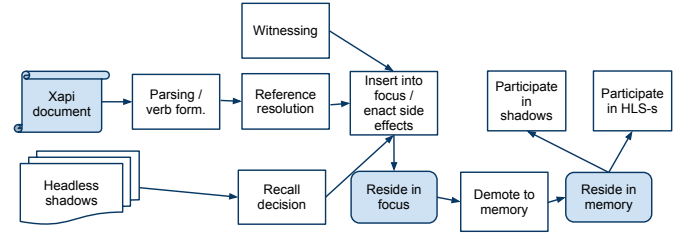

Fig. 4.    The life cycle of a verb instance.

### D. Verbs and verb overlays

The Xapagy system treats verbs very similarly to concepts.

A *verb* or *verb concept* in Xapagy is the representation of an *undivisible attribute* of an action or event. A weighted superposition of verbs is called a *verb overlay* (VO). When talking about verbs in general, we will denote them with $v_1, v_2 \ldots$ and VOs as $VC_1, VC_2 \ldots$. For specific verbs we will use descriptive names in brackets such as `[hits]`. The formulas introduced for COs will hold for VOs as well.

In contrast to concepts which are fully defined by their overlays and impacts, Xapagy defines a group of special *meta-verbs* which, when activated (like a verb), trigger *side-effects*, in the form of modifying the focus and creating relations between instances / verb instances. More detail about meta-verbs will be provided in Section V-D.

### E. Instances, attributes and relations

An instance is formed by a random identifier and the associated CO. Instances are created whenever a new entity appears in the ongoing narration, or whenever an existing entity was changed such that the additivity of the CO can not be maintained. The source of the instance can be external (when witnessing or reading a story) or internal (when recalling or confabulating a story). Figure 3 describes the life cycle of an instance.

In Xapi sentences, the creation of a new instance can be triggered by making a reference prefixed with an article a/an. This triggers the creation of a verb instance with the meta-verb `[create-instance]`, which creates the new instance and immediately initializes its CO with the attributes through which the reference has been made.

```
1   A man / exists.
```

The attributes of the concept can be changed in the future through `[is-a]` sentences, which enact an impacted addition ⊞:

```
1   The man / is-a / "Hector".
2   "Hector" / is-a / courageous.
```

The attributes of the instance are permanently attached to the identifier and they are *strictly additive*. Addition of new attributes is supported only to the extent that no significant negative impacts are exercised over the existing attributes. The only way we can turn Hector into a coward is through a `[change]` action:

```
1   "Hector" / changes / coward.
```

This operation, however, creates a new instance, which inherits a copy of the attributes of Hector, with the appropriate new concepts (and their impacts) applied. The old instance is immediately demoted to memory.

### F. Verb instances

A verb instance (VI) is composed of a random identifier and a number of mandatory relationships to its *parts*. Parts can be instances, VIs, COs or VOs. Table I describes the verb instance types and their mandatory parts. We can refer to the parts of a VI through a functional notation: for instance $SubI(V)$ is the subject instance of VI $V$. The only part shared by all VIs is the VerbVO(V), the *verbs*. The composition of this VO determines the *type* of the verb instance. For instance, the presence of the verb `[is-a]` implies a type of `S-ADJ`.

Figure 4 describes the life cycle of a VI. Most VIs are instantiated from Xapi statements. Those parts which refer to existing instances or VI are determined through the *reference resolution* process (see Section VII-B). The source of the verb instances is either external (from observation of ongoing events or reading/listening to a narration) or internal (from recall or confabulation).

## IV. FOCUS AND SHADOWS

### A. Terminology and notation

To simplify this presentation, let us start by several definitions. We define a *weighted set of instances $IS$* by saying that the participation of instance $I$ in instance set $IS$ is a value $w(IS, I) \in [0, 1]$. In contrast to COs, where concepts have impact and overlap, instance sets deal with independent and non-interacting instances.

The primary operation of instance sets is the *ratio-add*:

$$\forall r \in \mathbb{R} \quad IS = IS_1 + r \cdot IS_2 \Leftrightarrow$$
$$\forall I \quad w(IS, I) = trim\left(w(IS_1, I) + r \cdot w(IS_2, I)\right) \quad (9)$$

Verb instance sets *VS* are defined analogously.

The state of a Xapagy agent is modified by two kind of *activities*: *spike activities* (SA) and *diffusion activities* (DA).

SAs are instantaneous operations on overlays and weighted sets. Examples of activities modeled by SAs include inserting an instance in the focus, inserting a verb instance in the focus, and enacting the side effects of a verb instance. SAs are not parallel: the Xapagy agent executes a single SA at a time.

DAs represent gradual changes in the weighted sets; the output depends on the amount of time the diffusion was running. Multiple DAs run in parallel, reciprocally influencing each other. As a practical matter the Xapagy system implements DAs through sequential discrete simulation, with a temporal resolution an order of magnitude finer grained than the arrival rate of VIs.

### B. Story following and the focus

The focus is the collection of instances and VIs currently active in the Xapagy agent. Instances and VIs in the focus are *changeable*: they can acquire attributes, and they can enter into new relations. Once an instance or VI leaves the focus (is *demoted to the memory*) it will not change, except through a gradual decay.

The focus can be seen as two weighted sets of instances $IS_F$ and VIs $VS_F$ respectively. The focus is maintained through a number of SAs and DAs. In the following description we will list the SAs and DAs in an informal manner. The exact formulas are beyond the scope of this paper.

Let us start by listing the activities which affect the instances of the focus $IS_F$. We will prefix their descriptions with S and D depending on whether it is spike or a diffusion activity, and with +, - or +/- depending on their effects of the participations in $IS_F$.

(S+) instances are added at full participation at their creation.

(D-) in absence of other factors, the participation of instances decays in time.

(S+) instances referenced by newly created VIs are reinforced.

(D+) reinforcement on instances transfers part of the reinforcement to instances which are in a relation with the instance.

(D-) forced decay on instances transfers part of the decay to instances which are in relation to the decayed instance.

(S-) instances on which the change procedure has been performed are decayed to zero and removed from the focus.

The weighted set of VIs in the focus $VS_F$ represents, informally speaking, recent verb instances which are still relevant to the interpretation of current events. In general, the older a VI, the less likely to be still in focus. In practice, this is somewhat more complex than a gradually decaying tail of recent events, as the nature of the event changes the speed at which the VI fades from the focus. Non-action verbs, such as `is-a` will fade from the focus very quickly, action verbs tend to linger until they have acquired a sufficient number of successors while summarization verbs remain until they are a current summary and then fade out when they acquired sufficient successors.

Let us now summarize the activities which positively or negatively affect the participation of the verb instances in the focus:

(S+) VIs are added at full participation when created.

(D-) in absence of other factors, VIs decay at a speed dependent on their verbs.

(S-) VIs decay with every establishment of the successor relationship.

(S+) the participation of the summary verbs is increased by every matching VI added to the focus.

(S-) the `[is-single-scene]` verb decays all the current VIs from the focus.

When the participation of an instance or VI falls below a threshold, the instance or VI is removed from the focus. While there are several ways through which an instance might be reinforced in the focus, for a non-summary VI the only question is how quickly it will fade away. While instances and summarization VIs can (theoretically) stay indefinitely in the focus, non-summarization VIs will leave quite quickly.

### C. Shadows

Instances and VIs which have been demoted to memory can affect the current state of the agent by *shadowing* the focus. Each instance (or VI) in the focus is the *head* of a an associated instance set (or verb instance set) called the *body* of the shadow. The shadows are maintained such that their components reflect the previous experience of the agent with respect to the ongoing narration.

Shadows are dynamic and maintained by a number of activities:

(S+) The addition of an *unexpected* instance or VI creates a corresponding empty shadow.

(S+) The addition of an *expected* instance or VI creates a new shadow from the headless shadow which predicted the instance (for more detail see Section VIII-B).

(D-) In the absence of other factors, all the shadows decay in time. The energy released in this DA is added to the energy of the environment.

(D+) Matching the head: instances from memory which match the shadow head will be strengthened in the shadow body. The energy for this DA comes from the environment.

(D+) Matching the shadow body: instances from memory which match the shadow body will be strengthened in the shadow body. The energy for this DA comes from the environment.

(D+/-) Consistency: the participation of the VI in a shadow and the participation of its parts in the shadows of the corresponding parts of the shadow head gradually moves towards a common average value.

(D+/-) Instance identity sharpening: if an in-memory instance participates in multiple shadows, the strong participations will be gradually reinforced, while the weak participations will be further weakened. The operation is energy neutral for a given memory instance.

(D+/-) Non-identity: if a shadow contains instances which are connected through the non-identity relation[4], the instance with the stronger participation is reinforced while the instance with the weaker participation is weakened. The operation is energy neutral for a given non-identity pair.
instances which appear as different verb parts are guaranteed that they are not identical. This DA decreases the participation of weaker non-identical instances in the same shadow body. The operation is energy neutral based on a per / shadow body bases.

The shadow maintenance activities (and the closely related headless shadow maintenance activities controlling the narrative reasoning), are *self-regulating*, encompassing elements of negative feedback as well as resource limitation.

For instance, the head matching activity will not create an indefinitely large shadow even if the shadow head is very general, as the shadow instance set is "fed" from a limited set of resources, and once the shadow had grown beyond a certain size, its growth will slow.

Such interactions apply even between the activities. If a shadow is small, because there are few memory items matching its head, it can be extended by matching the shadow body, which brings in more remotely related items than the head match.

## V. CORE KNOWLEDGE

Problem domain knowledge appears in the Xapagy architecture in three forms:

- *Dictionary knowledge* includes the knowledge of the words in the Xapi language and the way they map to COs and VOs. Xapagy agents do not need domain specific grammatical knowledge, as the pidgin grammar is very simple and rigidly fixed.

---

[4]The non-identity relation is explicitly created for distinct instances in the same story line. For example, Achilles is non-identical to the instance of Hector with which it is currently fighting. However, Achilles is *not* non-identical with Lancelot.

- *Conceptual knowledge* is encoded in the properties of concepts and verbs. This includes their areas, overlaps, impacts and, for a small number of meta-verbs, the side effects of the verbs.
- *Episodic knowledge* covers the stories previously witnessed, read or confabulated by the agent. The stories are represented by the interconnected network of VIs and instances in the memory.

A Xapagy agent acquires dictionary and conceptual knowledge by loading knowledge description files in the Xapagy *domain knowledge language*. This very simple, human editable language allows the specification of the dictionary, the concepts and verbs together with their impacts and overlaps. The agent can save its dictionary and concept knowledge in the same format.

While the knowledge of Xapagy agents vary in all three domains (dictionary, conceptual and episodic), we assume that they share a common core of basic knowledge encoded in the Xapagy Domain Knowledge Library. The current presentation covers Xapagy CoreDKL version 0.20. We assume that the representations in this library are shared by all Xapagy agents. The CoreDKL only contains conceptual and dictionary knowledge. The agents share basic concepts and even elements of language, but they do not, by default, share remembrances of stories.

To reason about other problem domains, the agent can load other problem domain definitions. For instance, the `Iliad` library contains descriptions of words such as warrior or sword, as well as fighting related verbs. The `LittleRedRidingHood` domain contains definitions for hood, basket and wolf. Both domains rely upon and therefore load the `Human` library which provides definitions for words such as human, man, woman, alive and dead.

The reminder of this section discusses some of the functionality defined in the CoreDKL.

### A. Scenes

A scene in Xapagy is a *partition of the reference space*. Every Xapi sentence resolves its references inside a single scene, which needs to be resolved before the resolution starts. From this modus operandi follows that for VIs created from Xapi, only instances which are members of the same scene can interact with each other.

A scene does not necessarily correspond to a physical location, and the mechanical rules of interaction do not always apply. For instance, a conversation, either face-to-face or through remote means such as by phone, is a scene. If an agent talks to a friend in a restaurant, while being on the phone with another, it will be simultaneously part of three scenes: two non-physical conversation scenes, and the physical scene of the room, including the furniture, food, his friend and other customers.

Scenes are implemented as relations between the scene instance (an arbitrary instance which has the `[scene]` attribute) and the instances which are considered to be in the scene. An instance can be part of multiple scenes.

There is *current* scene in the focus. Newly created instances will be part of the current scene. Other scenes can be created as any other instance. The current instance is set with the `[is-current-scene]` verb. The statement:

```
1   A forest scene / is-current-scene.
```

creates a new scene with the attribute forest, and makes it the current scene. Previous scenes will be retained, although they will gradually expire from the focus. If we want to expire from the focus all the previous scenes, it can be done with the `[is-only-scene]` verb:

```
1   A forest scene / is-only-scene.
```

All the newly created instances will automatically become part of the current scene.

```
1   A forest scene / is-only-scene.
2   A "Little-Red-Riding-Hood" / is-a / girl.
3   The girl / has / a red hood.
4   The girl / has / a basket.
```

At this moment, the scene will contain the girl, the hood and the basket. An instance can leave the scene through the `[leave-scene]` verb:

```
1   The wolf / leaves-scene.
```

Instances which are not part of any scene but are still in the focus are considered to be part of the *limbo* scene. At reference resolution, the limbo scene is searched after the current scene, allowing it to move instances from one scene to another.

```
1   A house / is-current-scene.
2   The wolf / enters-scene.
```

Scenes determine the reference space where a particular reference is made:

- a reference to a scene is resolved among the scenes in the focus.
- a reference to an entering instance is resolved in the context of the limbo scene.
- a reference in a quote verb instance is resolved in the scene determined in the inquit.
- in all other cases, the reference is resolved in the context of the current scene.

### B. Groups

The concept of a group in Xapagy allows us to have VIs where the subject or the object is a collection of entities. Groups are implemented as instances which have the special attribute `[group]`, and they can be in a *composition* relation with a number of other instances. Instances can be part of more than one group and they can join and leave groups dynamically.

An *explicit group* is a group of individually known instances:

```
1   An explicit group / exists.
2   The group / contains / Billy.
3   The group / is-joined-by / "Johnny".
```

The side effects of most meta-verbs distribute over the members of explicit groups. For instance, we can remove all the members of the group from the current scene by saying:

```
1   The group / leaves-scene.
```

In the case of *implicit groups* the elements are not enumerated:

```
1   A group / exists.
2   The group / is-a / many man.
```

An implicit group can still have explicit members:

```
1   The group / contains / "Johnny".
```

At this moment, we have a crowd of people, out of which we recognize Johnny.

### C. Ownership-type relations

Let us call ownership-type relations the multitude of complex relations commonly expressed in English with the word "has". As Xapi does not support polysemantic words, the different relations must be represented in Xapagy by different words (and of course, different VOs).

The current version of core domain knowledge defines three distinct concepts `[holds]`, `[legally-owns]` and `[is-part-of]`[5], mapped to similarly named words:

```
1   "Little-Red-Riding-Hood" / holds / a basket.
2   "Mary" / legally-owns / a little lamb.
3   A big eye / is-part-of / the wolf.
```

---

[5]The `[contains]` verb of group membership discussed in the previous section is a close relative.

Ownership-type relations can change frequently as a result of sentences which mix in the VO the verb for the creation (or negation) of the relation, the meta-verbs for denoting an action verb and one or more verbs describing the nature of the action:

```
1   "Johnny" / holds / a ball.
2   "Johnny" / drops / the ball.
3   "Billy" / picks-up / the ball.
```

For instance, the word "drops" is mapped to `[drops not-holds succ*]`. Johnny can, of course get rid of the ball in a different way as well:

```
1   "Johnny" / puts-down / a ball.
```

where "puts-down" is mapped to `[drops not-holds succ*]`. Due to the limitations of the Xapi language (as well as the VI structure), the English statement "Billy gives the ball to Johnny" must be expressed as two Xapi statements (whose tight coupling is enforced by the "thus" word):

```
1   "Billy" / gives / the ball.
2   "Johnny" / thus receives / the ball.
```

One can consider this as the extreme case of the situation when Billy puts down the ball and, after a while, Johnny picks it up. This is an example of those cases where the limitations of the language can be addressed through a workaround involving instances shared among multiple VIs.

### D. Verb instance types as determined by meta-verbs

The side effects of a VI are determined by the *meta-verbs* participating in the VO their verb part. Although technically any combination of verbs and meta-verbs is possible, in practice most VIs can be classified in three classes, recognizable by their meta-verbs: *action* VIs, *attribute accumulation* VIs and *relation creation* VIs. These classes differ in their side effects as well as the amount of time they spend in the focus.

Action VIs represent actions or events which are part of the chain of a developing story. This is represented by the fact that they are connected by succession relations to their predecessors and by summarization relations to VIs which summarize them.

As action VIs need to be connected to their successors, they need to stay in the focus for a longer amount of time. Their participation in the focus is decreasing slowly, but it is reduced, in the form of a SA by the establishment of a succession relation: action VIs are *pushed out* by their successors.

There are two kinds of attribute accumulation VIs. One of them impacts instances by adding new attributes with the meta-verb `[is-a]`, while the other one impacts VIs by adding new verbs with the meta-verb `[action-is]`. Unless an attribute accumulation VI is also an action VI, it will expire very quickly from the focus (and in consequence, it will also have a low salience in the memory).

Relation-creation VIs (always of type `S-V-O`) embody a relation among instances in the focus. This type of VIs never expire from the focus as long as the represented instances are present. They can be, however, pushed out from the focus by other relation creation VIs.

Each verb in the VO can bring its own side effects which are implemented as a SA. The spike is normally triggered at the moment when the VI is inserted in the focus, although some type of insertions, such as for questions, inhibit the side effects.

### E. The succession chain

The succession relation is established between a *predecessor* VI and a *successor* VI when a new action VI is inserted into the focus. The SA activated at the insertion of verb instance $V_n$ creates succession relations relations between $V_n$ and all the VIs $V$ in the focus, the strength of the relation given by:

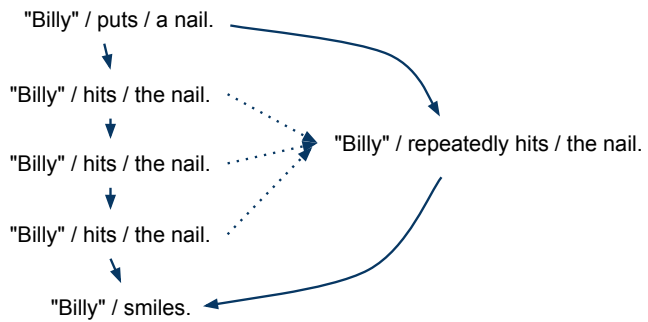

Fig. 5. Summarization of a repetitive action. Arrows with solid lines are successor relations, while arrows with interrupted lines are summarization relations.

$$r_s(V_n, V) = \sigma \cdot (w \cdot M(V_n, V) + 1 - w) \cdot focus(V)^r \quad (10)$$

The matching relation $M(V_n, V)$ takes the value 1 if $V$ and $V_n$ have at least one common part, and 0 otherwise. The balancing factor $w$ defines whether succession relations are established only with VIs which share parts with the new VI ($w = 1$), with all VIs in the focus ($w = 0$) or intermediary values.

The value $\sigma$ is the strength of SA. The parameter $r$ modulates the strength in function of the recentness of the links. If $r >> 1$ the link is connected only to the most recent VIs (for instance, in the case of the word "thus").

### F. Summarization relationships

Action statements represent individual actions or events whose timespan is measured over the course of seconds. Repetitive or continuous actions are *summarized* by summarization VIs.

Figure 5 shows an example of the summarization process (the VIs are illustrated with the corresponding Xapi statement). The left column of VIs are created by direct observation, while the VI in the right is created by the summarization process. The summarization VI is created when the second "hits" verb instance is issued. As long as the ongoing VIs inserted in the focus match the current summarization VIs, they will be linked to the current summarization VI.

Like other VIs, summarization VIs will participate in succession and summarization relations. This allows future recall processes to recall some stories in more details than others.

Multiple summarization VIs can be active simultaneously in the focus. Some of them might be summaries over VIs which are themselves summaries.

An additional complexity of summarization involves group actions. In the Figure 5 the summarization verb instance is simply a repetitive version of the ongoing verb instances which point to the same instances as subject and object.

However, we can summarize series of hits, kicks, blows and strikes exchanged between Hector and Achilles with a statement saying that "the two warriors were fighting". That is, the series of `S-V-O` verb instances are summarized by an `S-V` verb instance, there the subject part is a group instance.

There is only incipient support in the current version of the Xapagy system for this type of reasoning, thus the detailed discussion of the approach is beyond the scope of this paper.

### G. Creation of instances

The meta-verb `[create-instance]` creates a new instance with a specified set of attributes and adds a relationship that the new instance is a participant in the scene.

```
1  Scene / create-instance / attribute-list.
```

In most cases, this kind of statement does not need to be explicitly issued in Xapi, because it is automatically created as a side effect whenever an a/an article is used. The statement is created explicitly by the recall mechanism. If the recall is verbalized, the verbalization component might roll the instance creation into the sentence where the new instance is first referred to.

### H. Additive composition of attributes

An instance can acquire a new attribute by impact-adding to the CO associated with the instance, another CO described in `S-ADJ` type VI. This functionality is the side effect of the `[is-a]` verb:

```
1  "Hector" / is-a / warrior.
```

The general assumption about Xapagy instances is that they reflect a story entity with an unchanging set of attributes. The continuous addition of attributes through `[is-a]` is, thus, considered to be an act of *discovery*, not a recording of *change*. The added attributes should not have a negative impact on the existing set.

If a real change in the attributes is what we want to record, this must be done through the `[change]` verb, which creates a new instance).

Similar considerations apply to the VIs, whose verb part can acquired new attributes through an `S-ADV` type verb, as a side effect of the `[action-is]` verb:

```
1  "Achilles" / strikes / "Hector".
2  The strikes / action-is / hard.
```

### I. Creation of a new instance through change

If the story entity had changed its attributes (lost existing ones or acquired new ones) in a manner not consistent with the discovery of a previously unknown but existing attribute, the Xapagy agent must represent it through the creation of a new instance[6].

This functionality is implemented with the `[changes]` verb:

```
1  "Hector" / changes / dead.
```

a new instance is created, which acquires the previous attributes of Hector (including the proper noun "Hector"), to which the `[not-alive]` concept is impact added, triggering a number of negative and positive impacts (e.g. the removal of the `[alive]` concept from its attributes). At the same time, the initial instance is immediately retired from the focus and a relation of type `[is-somatically-identical]` is created between the old and new instances[7].

The side effects of the verb `[changes]` are described by the following SAs:

(S+) the creation of the new instance, with a set of attributes obtained by the impact-add from the attributes of the old instance

(S+) creation of the `[is-somatically-identical]` relation between the old instance and the new one

(S+) creation of scene membership relations to mirror the scene memberships of the existing instance

---

[6]As the creation of a new instance involves simply picking a random identifier from the existing flow, the creation of an instance does not incur any cost in the Xapagy model.

[7]Somatical identity, in the philosophy of personal identity represents the identity following the "body" of the entity. This is to be contrasted to the psychological identity which maintains the continuity of the memory (for entities for which this is relevant).

(S-) expire from the focus all the scene membership relations of the old instance

(S-) expire the old instance from the focus

### J. Destroying an instance

Certain real-world actions such as eating, drinking, or dissolving lead to the disappearance of that entity as an identifiable part of the scene. If this entity was mapped to a Xapagy instance, the instance can not be destroyed, as instances and their attributes are unchangeable. The real world destroying operation is echoed in the Xapagy agent through the removal of the instance from the focus and the scenes.

This can be implemented through the `[destroy]` meta-verb, which is part of the VOs associated with words such as "eats", "drinks" or "swallows".

```
1  "Billy" / eats / an apple.
```

On the other hand many operations which informally are denoted as destruction, are in the Xapagy parlance not a destroy operation but *change*:

```
1  "Billy" / drops / the bottle.
2  The bottle / changes / numerous-group glass shards.
```

### K. Representing questions

A question in Xapagy is a verb instance which is *partially unspecified* and it can form an *answer* relationship to VIs which match the question.

These VIs answering questions might already be in the focus, or they might be created on demand by the headless shadow mechanism. There is no requirement that there be a single answer to a question. Some questions might not be answered (not even attempted to be answered) or one might have multiple answers to the same question.

In general, the overall flow of the question / answer mechanism is beyond the scope of this paper. Here we shall only discuss the nature and specification of the question VIs.

The Xapagy CoreDKL defines a special verb `[wh]` and a corresponding special concept `[wh]` which, as part of a VO or CO respectively, turns the VI part into an unspecified one. Any VI which has at least one unspecified part is a *question*:

```
1  Johnny / wh cries?
2  Wh boy / cries?
3  Billy / hits / wh boy?
4  Wh / eats / "Little Red Riding Hood"?
5  Wh / wh / wh?
```

The `wh` marker marks the unspecified component of the question. All the verb instances marked above are questions, irregardless which component of the verb instance was marked with "wh". Once a VI has been marked as a question, it will inhibit the side effects of its verb (even if the verb itself is not marked with `wh`).

As the words in Xapagy map to VOs and COs, we can define specific words for frequently occurring combinations of `[wh]` and other concepts or verbs, such as "who" → `[wh human]`, or "where" → `[wh place]`.

### L. Representing a conversation

Xapagy defines as conversation the exchange of communicative acts between two entities in a story. One can represent an exchange of words without understanding the meaning of the communication. In this case, no additional knowledge is needed beyond the definition of the communicative acts themselves ("say", "ask" and so on).

If, however, we understand the content of the communication, we need a method to simultaneously represent the communicative act and the meaning of the communication.

The exhaustive discussion of the communication models is beyond the scope of this paper. We will illustrate the representation through one example, the famous exchange between the wolf and Little Red Riding Hood.

We are considering the external view of the story: that is, the case of a Xapagy agent reading the Grimm brothers translated to Xapi. The understanding proceeds from the point of view of the Xapagy agent: it does not require the wolf and the girl to be Xapagy agents.

The representation requires two scenes. The physical scene is the house of grandma - this is where the communcation will take place. The other scene, the *conversation* scene is a virtual scene where the meaning of the conversation is interpreted, and as such is marked with the `[conversation]` attribute.

The wolf and the girl are discussing about issues relevant to the current moment and themselves, thus they are part of both the physical and the conversation scene. This is not a requirement: if they would talk about the Iliad, the conversation scene would contain Hector and Achilles, but not the conversing parties.

The utterances of the conversation will be represented as quotation-type sentences, where the inquit is resolved in the physical scene, while the quote is resolved in the conversation scene. This being a quote sentence, the special resolution rules apply for "you", "me" and so on (see Section VII-B).

Putting it all together, the conversation between the girl and the wolf is translated to Xapi as follows:

```
1   A scene house / is-current-scene.
2   The wolf / enters-scene.
3   "Little-Red-Riding-Hood" / enters-scene.
4   A scene conversation / exists.
5   "Little-Red-Riding-Hood"/ enters / conversation.
6   The wolf / enters / conversation.
7   "Little-Red-Riding-Hood" / says / conversation //
8     Eyes / is-part-of / you.
9   "Little-Red-Riding-Hood" / asks / conversation //
10    Eyes / wh is-a / big?
11  Wolf / answers / conversation //
12    Eyes / see good / you.
13  "Little-Red-Riding-Hood" / says / conversation //
14    I / changes / afraid.
15  The wolf / swallows / "Little-Red-Riding-Hood".
```

### M. Representing a narration

A narration is a special case of conversation, which contains repetitive communicative acts by a narrator. The narration scene does not normally overlap with the current scene:

```
1   A scene "Smyrna" / is-current-scene.
2   A man / is-a / "Homer".
3   A scene "Iliad" / exists.
4   Scene "Iliad" / is-before / scene "Smyrna".
5   "Homer" / says / "Iliad" //
6     A man / is-a / "Achilles".
7   "Homer" / says / "Iliad" //
8     "Achilles" / is-a / angry.
```

In this sample, the man Homer, in his birthplace Smyrna starts the narration of the Iliad. Line 4 defines the temporal relationship between the scenes, i.e. establishing the fact that scene of Iliad happens in the past of the narration. The specification of such a relation is optional. Furthermore, it is possible to place a narration in the future. For instance, describing the prophecy made by Calchas we need to position it in the scene which is in the future:

```
1   Scene "Iliad" / is-before /
2     scene "ProphecyOfCalchas".
```

An agent can narrate a past (or future) story in which itself participates - but it will need to represent itself with another instance, with which it would establish some kind of identity relation (for instance, somatic identity). Let us see an example of Achilles narrating its own story at the Greek camp:

```
1   A greek camp scene / is-current-scene.
2   A man / is-a / "Achilles".
3   A scene "Remembrance" / is-before / scene camp.
4   "Achilles" / says / "Remembrance" //
5     A man / is-a / "Achilles".
6   "Achilles" / says / "Remembrance" //
7     I / somatically-identical / "Achilles".
8   "Achilles" / says / "Remembrance" //
9     "Achilles" / kills / "Hector".
```

## VI. Episodic knowledge

*After a minute or so, he goes out to join Doug, who is ritualistically lighting up a cigar. "This is a good time to smoke," he mumbles. "Want one?"*

*"Sure. Thanks." Randy pulls out a folding multipurpose tool and cuts the end from the cigar, a pretty impressive looking Cuban number. "Why do you say it's a good time to smoke?"*

*"To fix it in your memory. To mark it." Doug tears his gaze from the horizon and looks at Randy searchingly, almost beseeching him to understand. "This is one of the most important moments of your life. Nothing will ever be the same. We might get rich. We might get killed. We might just have an adventure, or learn something. But we have been changed."*

("Cryptonomicon" by Neal Stephenson)

Informally, the episodic knowledge of a Xapagy agent is the totality of the stories ever witnessed, read, heard, recalled or confabulated by the agent. Technically, the episodic memory is a repository of all VIs and instances which have been, at some time, part of the focus, and is implemented as two weighted sets (of instances and VIs).

The episodic memory is neither addressable nor directly searchable. The only way in which the content of the episodic memory influences future behavior of the Xapagy agent is through the shadow and headless shadow mechanisms.

We call the participation of the instances $S(I, t)$ and VIs $S(V, t)$ in the episodic memory their *salience* (at time $t$). The salience is maintained by two DAs:

(D+) Memorization - the salience of the instances and the VIs increases while the instance or VI is in the focus

(D-) Forgetting - the salience of instances and VIs exponentially decays in time.

### A. Memorization

The memorization DA increases the memory salience of an instance (or VI) during its stay in the focus. We will describe the equations for the case of VIs, the case of the instances is similar. Let us assume that the VI enters the focus at the creation time $t_c$ and leaves it the demotion time $t_d$. In between these times, it's salience will be:

$$S(V, t_x) = \int_{t_c}^{t_x} m(t) w(VS_F, V) dt \quad (11)$$

where $m(t)$ is the *marking rate* of the focus at time $t$. While the participation of the verb instance in the focus will gradually decrease, its salience will increase throughout its stay in the focus reaching its maximum salience at the moment when it is demoted to the memory:

$$S_{max}(V) = S(V, t_d) = \int_{t_c}^{t_d} m(t) w(VS_F, V) dt \quad (12)$$

What this formula tells us is that the memorization level is proportional with the time spent in the focus. Action VI are memorized more than attribute assignment VIs[8]. Action verbs are memorized better when they are not a part of a quick succession of events (when the successors quickly push out the VIs from the focus). Summarization VIs are remembered more strongly than individual events: the Xapagy agent might remember the repeated hammering of a nail, but not the individual act of hammering (see Subsection V-F).

In addition to this, the memorization also depends on the current marking rate of the focus. The marking rate is a slowly changing value, and it is *focus-wide*. Verb instances inserted in the focus affect the marking rate not only for themselves, but also for other verb instances before and after them. This way, it is possible that a marking action (such as the cigar smoking at the quote at the beginning of this section) would affect the memorization of an unrelated story line, which, however, share the focus with the marking action.

The current version of the Xapagy system uses a heuristic approach for the setting of the marking rate, which, however, appears to be successful in mimicking a wide range of behaviors. The formula is based on an exponential smoothing of the contributions of each inserted VI according to the formula:

$m'(t) = \lambda_m m(t) + (1 - \lambda_m) m_c(V)$

Currently, the smoothing factor $\lambda_m$ is set to 0.8. The continuation of the VI to the marking rate is dependent on the source of the VI. To allow for an explicit setting of the marking rate, the CoreDKL contains a special verb `[marker]` which allows us to artificially increase the marking rate. The currently used values for the marking rate contribution $m_c(V)$ are summarized in the following table:

| nature of $V$ | $m_c(V)$ |
| --- | --- |
| verb instances with the `[marker]` meta-verb | 1.0 |
| witnessed events | 0.5 |
| verbalized recall | 0.3 |
| non-verbalized recall | 0.1 |

Future versions of the Xapagy system will extend on this memorization model. Although it allows us to model explicit marker events, it does not model other aspects of human memory formation (such as the emotion caused by recalling certain experiences, which have not been very memorable when originally witnessed). There is a lot of existing work on human memory formation which can be modeled here - from the impact of levels of neurotransmitters in memorization (e.g. serotonin and dopamine levels), overall fatigue, as well as hard to measure aspects such as "interest".

### B. Forgetting

After being demoted to the memory, the salience of a VI decreases along an exponential decay curve:

$$S(V, t_x) = \lambda^{t_x - t_d} S_{max}(V) \qquad (13)$$

The salience of a VI will never increase after it leaves the focus. The recall of the VI does not increase its salience: it only creates a new, similar VI which might reinforce its recall (see Case 4 in Section VIII-C).

For most situations, however, the main challenge of adequate remembering is not the decreasing salience of the verb instances, but the initialization of the recall, which needs to create an appropriate focus and shadow.

---

[8]But the attribute itself is retained, because that one is dependent on the instance, which might have spent a lot more time in the focus. Thus the Xapagy agent might remember an attribute of the instance, but not when it acquired it - mimicking the limitations of human source memory.

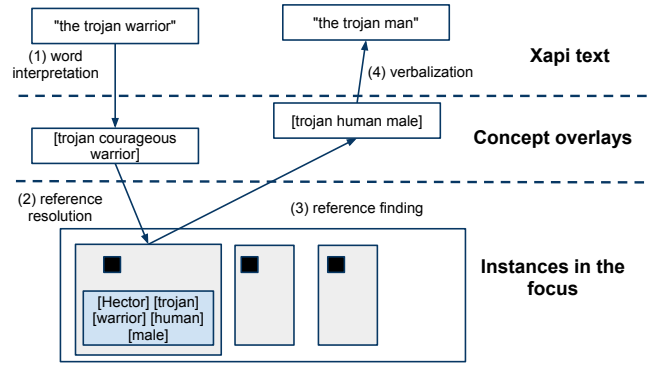

Fig. 6. The two levels of indirection between Xapi text and instances, and the four traversal steps.

### VII. THE XAPI ABSTRACTION AND REFERENCE RESOLUTION

The main external means of communication between a Xapagy agent and an external entity (let us say, a human person or another Xapagy agent) is the Xapi pidgin language.

When we refer to an internal element of a Xapagy agent (let us say, the instance of Hector) we need to pass through at least two steps of indirection. If we say "the trojan warrior", these *words* are first mapped into COs by a dictionary, then then two COs are merged together into a single reference CO describing the sentence part. Second, the reference CO must be mapped to an internal instance, through the reference resolution process[9]. When the agent communicates with the outside world, the indirection levels must be traversed in the opposite order, from instances, to concept overlays and on to words[10].

Figure 6 describes the four operations traversing the two indirection levels.

**Word interpretation** is the process which, starting from a list of one or more Xapi words forming a sentence part creates a reference CO (or VO) corresponding to these words.
**Reference resolution** is the process of finding in the focus the instance (or VI) to which the reference CO (or reference VO) refer to.
**Reference finding** is the process of creating a reference CO for a particular instance.
**Concept verbalization** is the process of finding one or more Xapi words, which wen interpreted, will approximate a specific reference CO.

The activities on the upward path (reference finding, and concept verbalization) are always performed with the expectation that the generated Xapi text will traverse a downward path in a target agent. The target agent might be identical to the source agent, in case of self-narration. But most typically it is a different agent.

This opens an apparently impossible challenge: the target agent might have a different dictionary and different concepts. In the strong philosophical sense there is the impossibility of perfect understanding

---

[9]It is quite possible that future developments in the Xapagy architecture, such as the introduction of word polymorphism would increase the number of indirection levels.

[10]Note that not every conceivable concept overlay can be expressed using words, especially if the agent has a limited dictionary. We might be tempted to declare this an example of linguistic relativity, in the interpretation of the Sapir-Whorf hypothesis and its modern revisions. But as here we are talking about a pidgin language artificially designed to have a low representational power, drawing any further conclusions would be circular reasoning.

[16], but even in a purely technical reading there is a major challenge that an agent can not generate an optimal output unless it knows the dictionary and concepts of the target agents. Although we are not oblivious to these theoretical and conceptual difficulties, in practice we have adopted a solution which is not very different from what a human might do: perform the upward path with the assumption that he itself is the target, add a safety margin to the reference finding, and, if necessary, restrict the vocabulary to match the vocabulary of the target agent.

Throughout our discussion below, we will consider the target of the reference to be an instance. Reference to VIs occurs comparatively rarely (in adverbial sentences only) and it is largely analogous to the instance case.

### A. Word interpretation

Xapi sentence parts are composed of one or more words. These are mapped into COs or VOs by the word interpretation process, some of them being further processed as references to an instance.

For the generated COs and VOs we will use the name *reference CO/VO* if it is used as a reference, *template CO/VO* if it is used as a template to create a new instance or VI, and *attribute CO/VO* if it is used to change the attributes of an instance or VI. The word interpretation process, however, is identical for all these cases. The following table shows the ways in which different sentence components are interpreted and how there are going to be further processed.

| Sentence part | Interpreted as | Next step |
| --- | --- | --- |
| Verb | template VO | VI creation |
| *the* SubI | reference CO | reference resolution |
| *a/an* SubI | template CO | instance creation |
| *the* ObjI | reference CO | reference resolution |
| *a/an* ObjI | template CO | instance creation |
| SubVI | reference VO | VI reference resolution |
| ObjCO | attribute CO | impact-add as attributes |
| ObjVO | attribute CO | impact-add as VI attributes |

For example, in the sentence:

```
1   The tall red man / is-a / angry.
```

the subject part "the tall red man" is a SubI, so will be first mapped to a reference CO of the specified attributes, then the reference resolved to find an instance which matches those attributes. The ObjCO part "angry" will be similarly mapped to an attribute CO, but it will not be further resolved - this CO will be used to add to the subject instance, as a side effect of the verb `is-a`. The word "is-a" will be interpreted as a template VO and will be used to create the VI for this sentence.

In the following, we will present a discussion of the challenges of the word resolution, and the current set of compromises and solutions adopted by the Xapagy system.

The current version of the Xapagy system performs word interpretation by direct dictionary lookup followed by *cross referenced impacted addition*. The dictionary lookup in the current version is context independent[11] - the same word will create the same CO independently of which sentence part it occurs in or the current content of the focus and shadows.

Let us explain the rationality behind the *cross referenced impacted addition* method. Naturally, the simplest solution would be to simply sum the COs mapped to the different words in the dictionary. The

---

[11]More exactly, the Xapagy system has in place the architecture to support context dependency: however, in our current work, we did not found the need or the representational advantages for it.

---

problem appears when these overlays have elements which have negative impacts with each other.

Let us see an example. The word `warrior`, maps to a CO including the concept [`human`], which in its turn, impacts [`alive`]. So if we have a sentence such as:

```
1   A warrior / exists.
```

what the agent will model is a warrior who is alive. If we say however:

```
1   The dead warrior / is-a / "Hector".
```

what we mean is a dead warrior. We do not want the impacts to cancel each other out: the [`alive`] impact of the `warrior` word needs to be eliminated before it even appears on the overlay.

### B. Reference resolution

Reference resolution finds an instance in the focus which, in a way, *matches* the reference CO. In Xapagy, the resolution process has two steps:

- determine in which scene the reference resolution would take place
- determine which instance in the specified scene is the target of the reference

**Scene resolution:** the agent maintains a current scene which is set explicitly by the [`is-current-scene`] and [`is-only-scene`] the verbs. The current scene is used for the resolution of all primary sentences.

For the quote component of the quotation sentences, the scene used to resolve the references is the embedded sentence of the instance is obtained from the ObjI part in the inquit, which first, in itself, needs to be resolved within a virtual scene composed of the other scenes. **Instance resolution:** in the Xapagy system is based on the *elimination of the alternatives*. The reference CO must be chosen in such a way as to mismatch all the candidate instances which it does not refer to. The reference will be resolved to the strongest instance in the focus which it does not mismatch.

A reference CO $C_R$ is mismatched with an instance $I$ if:

- it has a highly specific concept not present in the attributes of $I$
- it has concepts which have a negative impact on the attributes of $I$

The mismatch between $C_R$ and the attributes $C$ of instance $I$ can be calculated with the following formula:

$$
\begin{aligned}
\mathit{mismatch}&(C_R, C) = \\
&\max_{c \in C_R} \left( \frac{en(c, C_R) - w \cdot en(c, C)}{area(c)} \right) \\
&\quad - \sum_{c \in C_R} min\big(impact(c, C), 0\big)
\end{aligned}
\tag{14}
$$

The instances which have a mismatch higher than a predefined threshold will be eliminated from the reference competition. It is not necessary to reduce the field of candidates to zero: if the field still has several candidates, the one with the highest current participation in the focus will be chosen as the referred instance. This technique allows for recentness based reference resolution:

```
1   "Achilles" / is-a / warrior.
2   He / is-a / trojan.
```

where the word "he" is dictionary mapped to [`human male`]. [`human male`] will not mismatch with neither Achilles, nor Hector or Ulysses. In the current context however, the newly referred instance Achilles will be the strongest instance in the focus, thus the reference will be resolved to him.

Such a reference becomes more and more ambiguous as the distance between the last reference to Achilles and the use of the word "he" is increasing. When we use the reference "Achilles", however, we use the highly specific concept [``Achilles''] which is only contained by the instance Achilles. This reference is unambiguous, because it will leave only one candidate (as long as there is only one person named Achilles in the story - for Ajax however, we need to use "great Ajax" and "lesser Ajax").

An interesting side-effect of the mismatch function allows *reference through negation*:

```
1    The not-trojan warrior / hits / "Hector".
```

*1) Special resolution rules for quotations:* As we have seen, the quote statements trigger special resolution rules. The elements of the inquit statement are resolved in the current scene, while the parts of the quote are resolved in the scene specified in the inquit.

In quoted statements there are some special resolution rules corresponding to pronouns. These are special words, which, despite the fact that they are in the quote, they resolve in the scene of the inquit:

- [I] or [me] in the quote resolves back to the inquit subject.
- [Thou] in the quote resolves to the strongest instance in the scene which is not group nor the inquit subject.
- [You] in the quote resolves to the strongest instance in the scene which is not the inquit subject (it can be a group).
- [We] in the quote resolves to the strongest group in the scene which includes the inquit subject.

### C. Reference finding

Reference finding is the process of finding a reference CO $C_R$ for a specific instance I. As we discussed in the introduction of this section, the Xapagy system assumes that the target agent to be the same as the source agent. However, additional criteria apply to the choice of the reference, and this criteria normally includes a preference against ambiguity in reference[12]. One way to trivially achieve a reference is to refer to instances with their complete set of attributes:

```
1    The greek warrior man "Achilles" / hits /
2      the trojan warrior man "Hector".
```

A better choice is to find a minimal reference CO, that is, an overlay which has the smallest energy but which eliminates all the other alternatives. As the proper names have the smallest area, this almost always means referring by proper name:

```
1    "Achilles" / hits / "Hector".
```

This reference mode, repeated over the complete course of the story appears boring to a human reader. The current implementation allows us to install a preference model for the chosen references. This can be used to implement:

- prefer the use of pronouns (he, she, it) whenever they resolve correctly
- avoid to refer with anything other than the pronoun I, you and thou in all occasions where these can be applied (i.e. the agent should not talk about itself in third person).
- the Homeric approach of using of a proper name with a returning epithet: swift-footed Achilles, man-killing Hector.
- preference for a certain type of attributes (e.g. nationality, age and so on).

However, the current version of Xapagy does not contain support for automatic setting or adjustment of these values.

---

[12]We choose either references which mismatch against all the instances in the scene except the referred item, or find a large gap between the relative focus participation of the referred instance and other candidates.

### D. Verbalization

The last step of our process involves the problem of selecting a collection of words from the dictionary such that they express the reference CO. Not all the possible COs can be expressed with words. If the CO we try to express is referring to an instance whose attributes have been acquired from Xapi, then, necessarily, the complete set of attributes can be expressed with words in the dictionary. But even than, the concepts chosen in the reference CO might not be expressible with words. Let us consider an agent which does not have a word for "male", and let us assume that the reference finding would choose [male trojan] as a way to refer to Hector. The agent might try to approximate this with "trojan man", but that is overreaching, as it will be interpreted as [male human trojan][13].

The current approach is to generate a series of candidate word combinations (limited to a maximum of two words) and then calculate the distance between the CO generated by these through word interpretation and the original CO we want to express. The distance calculation is based on two penalty values:

- the penalty for the unexpressed content (the part of the overlay which is not conveyed by words) $P_{UC} = f(C_{intended} - C_{words})$
- the penalty for the additional content (the additional concept conveyed by the words but not part of the recalled overlay) $P_{AC} = f(C_{words} - C_{intended})$

The set of words to be chosen are such that they minimize $P_{UC} + \alpha P_{AC}$ where $\alpha$ is the unintended meaning amplification ratio (which can be chosen freely but in our experiments we set it to 10.0).

## VIII. NARRATIVE REASONING IN XAPAGY

It is finally time to integrate the components discussed throughout this paper and describe how the architecture of the Xapagy system enables narrative reasoning. We start by describing the common framework of all narrative reasoning techniques, the *headless shadows* (Subsection VIII-A).

Then, we dedicate a subsection each to the following aspects of narrative reasoning techniques:

- **Story following** (Subsection VIII-B): predicting ongoing events, expressing surprise about events and filling in missing events.
- **Story creation** (Subsection VIII-C): creating new stories through recall or confabulation (or various shades in between).
- **Divagations** (Subsection VIII-D): creating VIs which do not advance the story line but can be necessary (such as the introduction of instances) or helpful (such as elaborating on instances or justifying VIs).
- **Story steering** (Subsection VIII-E) the means through which the story creation process can be guided either by the agent itself or external agents. This includes the recall or confabulation initiation process, the use of questions, as well as, in the extreme case, of collaborative story telling.

### A. Headless shadows

All the narrative reasoning techniques rely on the mechanism of *headless shadows* (HLS), collections of related and aligned in-memory VIs which are not paired with any current in-focus VI. Like shadows, HLS-s are maintained on an ongoing basis through a collection of SAs and DAs. All the narrative reasoning models can be understood in terms of a single procedural pattern:

- **maintain** a collection of HLSs reflecting the current state of the narration

---

[13]It is obvious that the reference finding and concept verbalization components will need to communicate in some way. This communication does not exist in the current version of Xapagy.

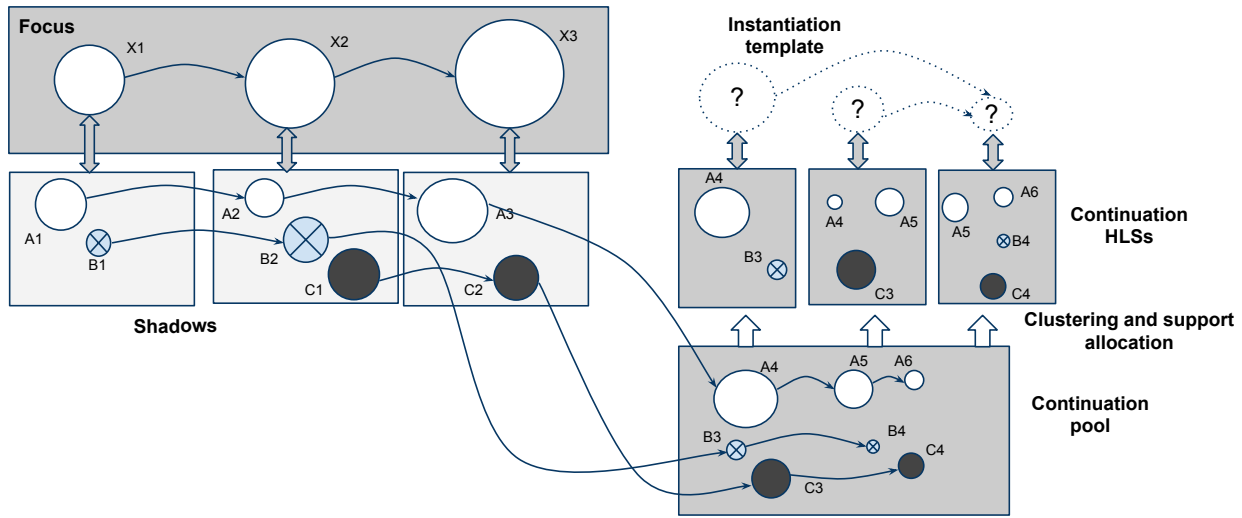

Fig. 7.   A simplified representation of the continuation HLS formation.

- **choose** an HLS for instantiation based on a specific criteria
- **instantiate** the HLS by creating a new VI
- **insert** the new VI to the focus and transform the HLS into its regular shadow
- **verbalize** the VI

The difference between the narrative reasoning methods stems from whether they skip one or more steps of the procedure, as well as the different criteria they might use to choose the HLS for instantiation.

HLSs can be created in several ways. The most direct model is the *continuation HLS*, which are created by the clustering of the successors of verb instances in regular shadows.

For illustrative purposes we will describe the HLS formation as a step-by-step process involving discrete steps. In practice, this is a continuous, ongoing activity performed by DAs.

Let us consider the example described in Figure 7. In the first step, the agent maintains three VIs in the focus, with their respective shadows (the rightmost VI is the most recent one). The size of the disk denotes the participation of a given verb instance in the focus or shadow, respectively. The focus is shadowed by VIs from three different stories. The VIs linked by successor relationships from the shadows create a *continuation pool*.

In the second step, the elements of the continuation pool are clustered in continuation HLSs. The HLS contains a *template* of the possible verb instance it can instantiate. VIs from the continuation pool *lend support* to certain HLSs based on their match with the template.

Many of the DA's for the maintenance of shadows also operate for HLSs (for instance, story consistency). However, the low level details of the SAs and DAs performing the HLS maintenance, as well as the discussion of boundary cases such as what happens if the shadow predicts the apparition of a new instance, is beyond the scope of this paper.

### B. Story following: Predicting ongoing events, expectation fulfillment and surprise

In this simplest narrative reasoning method the agent is a passive observer of the ongoing flow of events.

When a new VI is added to the focus from external sources (witnessing, reading, conversation), it will be matched against the continuation HLSs. If there is a match, the HLS will become the shadow of the new VI.

This activity is an SA, and in its turn will trigger changes both in the other shadows and the continuation HLSs. The story consistency DA, for instance, will ensure that the shadow components which predicted the right continuation will be reinforced and HLSs which are compatible with the new HLS will be reinforced. New HLSs are calculated starting from the current focus and shadow, while those which became incompatible with the current status will be lowered in support or even discarded.

We introduce two metrics with regards to the agent's perception of the incoming event: *expectation fulfillment* and *surprise*.

- The *expectation fulfillment* is the support of the HLS matched to the incoming VI (or zero is no such HLS has been found). If there is no match, this value is zero. We say that an event was *expected* if there was a HLS with a strong support matching it, and *unexpected* otherwise.
- The *surprise* is defined as the magnitude of change in the continuation pool, triggered by the event (according to an appropriate distance metric defined in the continuation pool). With this definition, even the least surprising event will have a non-zero value, as the continuation HLSs will adjust, extending themselves into the future. For purposes of analysis, we can define the *normalized surprise*, which is the surprise of the actually happened event minus the minimum possible surprise.

Expectation fulfillment and surprise are related metrics, but they can not be trivially transformed into each other. For many situations, the most expected event will generate the least surprise. However, we can have highly unexpected events, which do not create a surprise. Let us consider a scenario in which Hector, in the heat of the battle, stops to wipe his brow. This is unexpected: none of the HLSs could have predicted this. However, it has little impact on the predictions, so it is not surprising. Surprising events trigger a change in the continuations, as they determine the future flow of the story. The death of the Hector is not unexpected in the context of a duel. It has however a large surprise value, as it causes the Xapagy agent to completely change the continuation pool, as the continuations involving further actions by Hector will be removed.

### C. Story creation: recalling and confabulating

Recalling a story in Xapagy implies maintaining the HLSs, selecting an appropriate continuation HLS, instantiating it and inserting it into the focus. We call this operation a *recall* if the sequence of instantiated VIs closely resemble a previously experienced story. We call it *confabulation* if the new story does not resemble any previously declared story. Perfect recall can be achieved in the Xapagy system only under very strict, not naturally occurring conditions. On the other hand, there are many fine gradations between perfect recall and freely associative confabulations, most of these mimicking, in a recognizable way, human behaviors.

We will illustrate the range of possible recall / confabulation behaviors through a series of cases. These cases assume that the agent is already in recall mode, and the appropriate initalizations have been performed.

Due to space limitations, we need to describe each of these cases in a cursory manner. Their presentation, however, is necessary to understand the range of functionality which can be achieved through the recall mechanism of the Xapagy system.

To illustrate these cases, we will rely on a series of graphs. *These graphs are simplifications, they are not outputs of actual runs.* For the reader to get a good feel of what is presented and what is suppressed in these graphs, we need some introductory definitions.

We call a *story line* a collection of VIs which have been directly connected through succession links. One can, of course, envision the lifespan of the Xapagy agent as a whole continuous story line. In practice, however, long temporal spans, the replacement of the instances and scenes segments the agent's episodic memory into distincts episodes. Note, however, that the story lines are just an *ascriptive* definition of certain subsets of the episodic memory, which we make for easier readability (in this case, with graphs). The VIs are *not* labeled with story line identifiers.

To define a recall scenario, we assume that a desire to reproduce a specific story line from the past, which we will call the *dominant story line*. Other story lines which enter into consideration during the recall are called *foreign story lines*. As we move from simple recall towards confabulation, the difference between the dominant and foreign story lines become increasingly blurred.

Let us now define the format of the figuress we shall use. The figures describe the continuation HLSs, with their instantiation template (as a circle drawn with dotted line). The size of the instantiation template circle illustrates the total support of the HLS. The HLSs are represented by dark gray rectangles. Inside the rectangles, we have disks representing support from VIs in the episodic memory: the size of the disks illustrates the strength of the support. Some in-memory VIs can support more than one HLS; we only show them in the HLS which they support the most strongly. Support VIs coming from different story lines will be colored or marked differently. VIs marked with solid colors have been created from external sources (reading, witnessing, conversation), those marked with a symbol ($\oplus$ or $\otimes$) from internal sources (recall or confabulation). The dominant story line will be denoted with white disks. We show the succession links between the support VIs with arrows. In practice, a VI can have several succession links, we only show the strongest one.

Some figures will also contain the VIs in the focus, with their shadows. We do not represent instances in the focus and nor their shadows.

Figure 8 illustrates all these drawing conventions.

*1) Case 1: Pure recall of a witnessed story:* Pure recall is the accurate reproduction of the events in a story line, un-contaminated by the influence of other story lines in the episodic memory. This can happen if each HLS is supported by a single VI, from the dominant story line (we call this the *purity condition*).

For a practical agent, the purity condition is difficult to achieve. The "matching the head" and "matching the body" DAs will bring in shadows from any previous VI with a similar VO. The more experience an agent has in the specific domain, the more likely is that its shadows during recall will contain elements outside the dominant line. These shadow elements will send their successors into the continuation pool, and some of them will support some of the HLSs.

The purity condition can be maintained if either:

- the agent is completely clueless in the domain (it has no previous experience with any of the the concepts appearing in the recalled story line)[14].
- the story is so surprising that none of the other remembered stories can support the recall

Figure 9 shows the pure recall scenario. The HLSs, which are selected and instantiated from left-to-right, exactly mirror the VIs in the original story line. The purity condition is satisfied as each HLS is only supported by a VI from the recalled story line.

We emphasize that the purity condition applies only to the HLSs. The shadows and even the continuation pool *can* have some elements outside the dominant story line, as long as they do not reach the threshold to enter the HLSs.

*2) Case 2: Competitive recall:* As we have seen, the purity condition can be satisfied only under very restrictive conditions. In addition to the dominant story line a number of foreign story lines will (a) provide support to the HLSs generated by the dominant story line and/or (b) create *foreign* HLS-s, which are not supported by the dominant story line.

We say that the dominant story line *competes* for the recall with the other story lines. The dominant story *wins* the competition if the resulting recall is identical to the one from a pure recall. The dominant story *looses* to the other stories every time the recall is distorted from the pure recall. For Case 2 we assume that these losses are localized: the recalled story will still largely follow the dominant story line.

The dominant story can loose out to the foreign story lines in the following ways:

**HLS distortion:** happens if an HLS, created by the dominant story line is supported by one or more foreign VIs.

Let us assume that the agent tries to recall Homer's Iliad, where Hector was killed with a spear to his throat. But the agent has also seen the Hollywood movie Troy where Hector is killed with a sword to the chest. These two events are close enough to support the same HLS, but the different supports are not identical and the recalled verb instance might be different from the one in the dominant story (the Iliad in this case).

**Skipping story elements:** Let us assume that the dominant story line had generated four HLSs, which in a pure recall would be recalled in the order A → B → C → D (see Figure 10). However, B and D are strongly supported by a foreign story line, while C is not. Under these conditions, it is possible that the recall process would skik the poorly supported C, making the recall A → B → D.

---

[14]We must mention here that in the current version of the Xapagy agent it is possible to have conceptual knowledge without associated episodic knowledge. It is possible, for instance, for an agent to have a model for the verb "kill" without having witnessed, heard, or read about any story involving killing.

An of course, in philosophy, psychology and linguistics the existence of concepts predating experience is about the largest cans of worms which one can open.

It is not our intention to make a statement with this. In the current version of the Xapagy system this state of affairs is simply an artifact of the fact that we do not yet have in place a mechanism for concept learning.

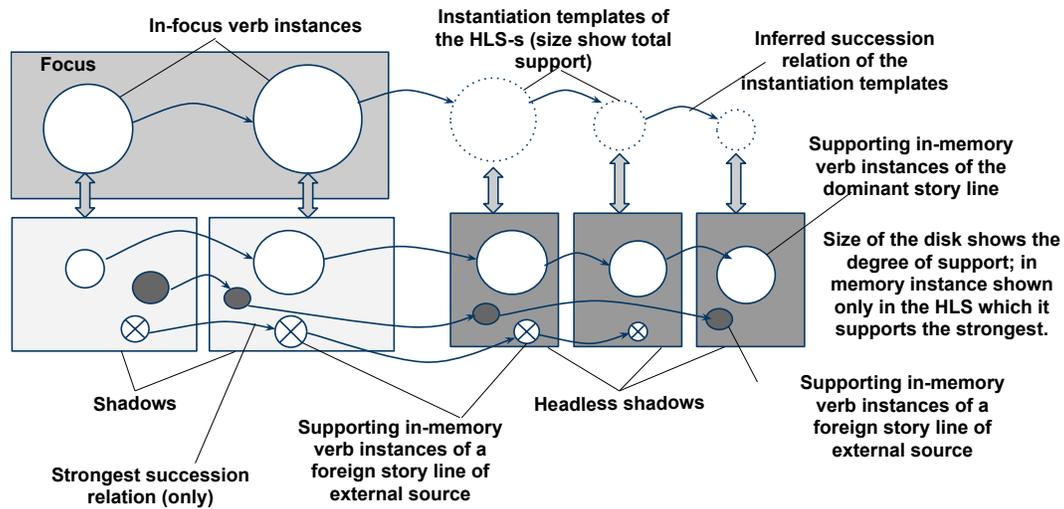

Fig. 8. Illustrating the conventions of the representations for the cases on the recall / confabulation axis.

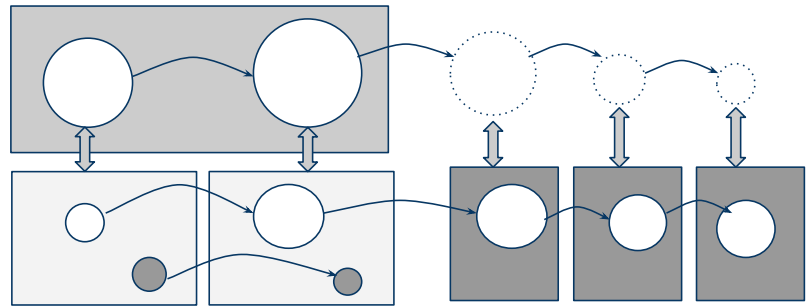

Fig. 9. Case 1: Pure recall of a witnessed story. The purity condition is satisfied: all the support of the HLSs comes exclusively from the dominant story line. The shadows, however, can contain components from the foreign story lines.

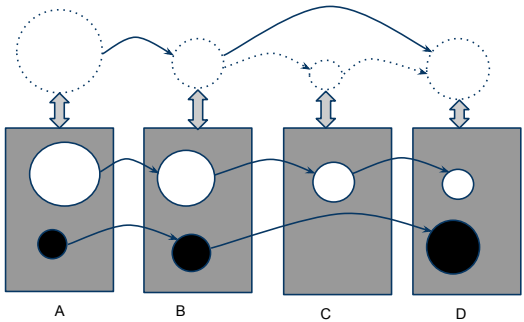

Fig. 10. Case 2: Competitive recall. The competing story line triggers the skipping of the story element C from the dominant story line.

**Inserting foreign story elements:** similarly to the previous case, the recall inserts a story element not present in the dominant story line, but strongly supported by one or more foreign story lines.

*3) Case 3: Recalling frequent events:* This case considers the challenge to recall one specific dominant story line from a collection of near identical story lines, such are recalling what one had for breakfast on a certain day. The psychology literature on autobio-graphical memory frequently calls this type of memories *repisodes*. The phenomena has been identified and well documented since the 1980s [8] and still under active research [4]. This is a well known challenge for the human memory, as the similarity between the events makes it more difficult to recall one particular event in an individual way.

Let us assume that the agent experienced a number of "breakfast" stories, involving either "bread-eggs-coffee" or "bread-sausage-coffee" type of breakfasts. Naturally, these story lines do not float in the empty space: they are embedded in the particular story line of the given day which includes what happened before, during and after the breakfast in the specific day.

The support structure of the HLSs this case is shown in Figure 11. HLSs for each typical breakfast action will be created. Each will be supported by large number of VIs, due to the large number of near-identical story lines. There will be little difference between their relative strengths and each of the supports will be small (due to the negative feedback mechanism in the match the shadow head DA). Thus, we have a variation on the competitive recall, with many weak, evenly matched competitors.

Accurate recall under these conditions can only be achieved if (a) there were some differentiating factors for the specific episode and (b) these differentiating factors can be recreated during recall. The flip side of this, however, is that truly repetitive activities, with no distinguishing factors, can not be accurately recalled for specific cases. If an agent had alternatively eaten sausage and eggs, with no other distinguishing factors in the story of the morning, it will *not* be able to recall what it did exactly in a specific day.

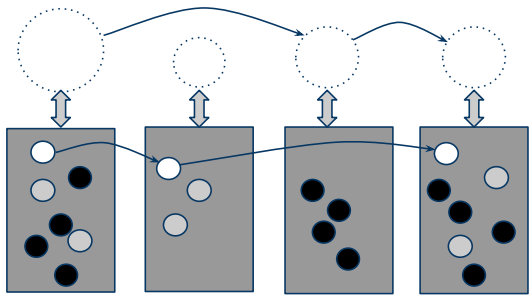

Fig. 11. Case 3: Recalling frequent events. Black: bread / sausage / coffee, White: break / eggs / coffee.

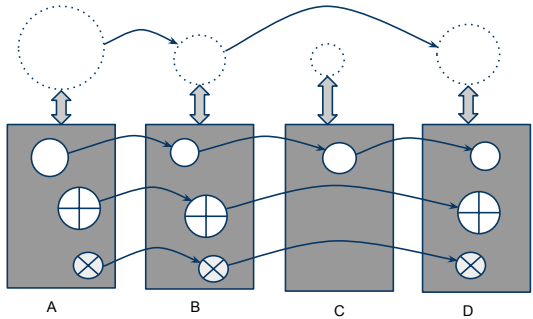

Fig. 12. Case 4: Self shadowing and drifting. The frequent ommission of the VI C from the retelling (for instance by constraints) drifts the story line, such that the retelling will proceed on the story line A → B → D even when no constraints are present.

*4) Case 4: Self shadowing recall. Drift:* In the previous case, we considered the case where the agent had witnessed many highly similar stories. This case deals with a variant of this case, where the agent had witnessed a story only once (and the story can be highly specific), but recalled it many times.

The Xapagy system differentiates between direct experience, hearsay and recall through the circumstances of an event (the scene, predecessors, successor, whether it is the quote part of a quotation statement). We can, for instance recall seeing a movie where Homer is narrating the Iliad. The story of the Iliad will be encoded through a two level quoting. Nevertheless, even in this case, the narrated Iliad forms a valid story, with the succession links chaining through the elements of the Iliad's story line (with separate links chaining through Homer's narration, and our watching actions, respectively).

These narrations are clearly separated: it is not the case that the Xapagy system cannot recall the source of the information provided that it is correctly initialized to recall the watching scene[15].

Despite the ability to explicitly recall the source of a specific story, the foreign story lines brought in by the DAs maintaining shadows and the HLSs will contain all types of verb instances both externally and internally generated, directly witnessed, read or heard. This is a necessary and useful part of the system - for example, allows a

Xapagy agent to make correct predictions when witnessing real world situations about which previously it had only "book knowledge".

The phenomena of self-shadowing and story drift is an inevitable consequence of this otherwise useful functionality.

Figure 12 illustrates the situation. The agent tries to recall the dominant story, which, we assume, was directly witnessed. The previous recalls are marked with ⊕ and ⊗ symbols. As these recalls are necessarily very close to the original story, they will inevitably support the specific HLSs. We call this phenomena *self-shadowing*.

This can have both positive and negative consequences. On the positive side, accurate recalls reinforce the HLSs of the dominant story, and make it less likely that foreign stories can compete with the recall. This is especially important if the agent recalls events for which the salience had naturally decayed.

With this mechanism, the Xapagy system automatically exhibits learning through repetition (it is a future task to investigate whether such observations about human learning such as the Eberhardt learning curve or Paul Pimsleur's spaced repetition technique can be automatically mimicked by the Xapagy agent).

Let us now consider, however, the case when the recalls have not been fully accurate. If the recalls differ from the dominant story line in a consistent way (for instance, by regularly skipping some events, the situation in Figure 12) there will be a strong likelihood that a specific recall will follow not the original story line but the "usual way of recalling it". This phenomena, which we will call story drift, mimics suppressed memories and self-deception in humans. A closely related situation is when the shadowing stories are not coming from internal recalls, but from external retelling of the same story in modified form. This might mimic human behavior where, for instance, excessive external praise might modify the person's own recollection of certain stories in the past.

One would hope, for instance that humans can simply filter out internally generated stories to affect the recall - but apparently humans do not have a build in filter for this purpose.

A very famous example of self shadowing is the case of Nixon's counsel John Dean analyzed by Neisser [18]. In several cases where Dean had recalled with high confidence details of meetings with Nixon, it turns out that what he had recalled was heavily modified by his fantasies and modified recalls of the event (both omission of events and insertion of events have been noted). It was also found that at its testimony, the main source of rememberence is not the original events but the statement about it he have a couple of days before.

*5) Case 5: Constrained recall:* The cases up to this point made the assumption that the goal is the accurate recall of the dominant story line. Case 5 deals with the case when the objective is to recall the dominant story line subject to a number of constraints. One possible reason to do this is to adapt the story to the preferences of the audience. We can use constraints to produce a bowdlerized version of a story, such as a children's version of the Iliad would not have Achilles kill Hector, only "defeat" him.

The current version of the Xapagy system implements constraints by adding an additional filtering layer, which changes the overall support of the HLSs[16].

If we are in a situation of pure recall, the only possible impact of the constraints is to remove certain events from the story line. In the case of competitive recall, however, the introduction of constraints can lead to the replacement or even insertion of new VIs in the recall as it locally changes the balance of power between the dominant and the competing foreign story lines.

---

[15] Although an accurate recall for the whole chain might be difficult to achieve, as the watching television activity most definitely fall under the category of "repisodes", creating the situation in Case 3.

[16] Future versions might operate at the level of individual supports inside the HLSs

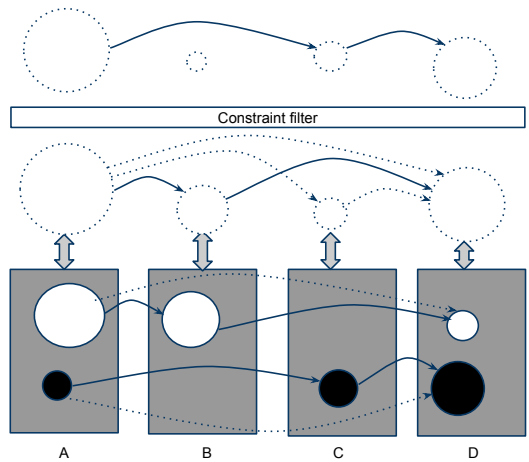

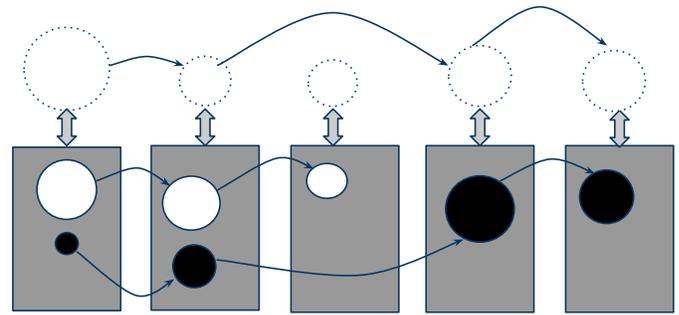

Fig. 15. Story line jumping. The VIs marked with black gradually increase in support and become the dominant story line.

Fig. 13. Case 5: Constrained recall. An example of a constrained recall inserting a foreign component in the dominant story line. The constrained recall will be A → C → D instead of A → B → D.

Figure 13 shows such an example. At the moment represented in the figure, 4 continuation HLSs are maintained. The chain of a future recall will be A → B → D, correctly following the dominant story line. However, the design of the constraint filter removes A from consideration, which brings into the recall HLS B, supported only by the foreign story line, which would not have been considered otherwise. After this divergence, however, the recalled story returns to the dominant story line, so the recall will be A → C → D.

*6) Case 6: Leading questions:* Case 6 deals with the situation where the recall deviates from the dominant story line in response to specific statements coming from an external agent. We call this case "leading questions" due to the obvious similarity to the influencing of the recall through leading questions as it might happen in a courtroom. However, the external text does not necessarily have to be a question.

For an external statement to affect the recall, it is necessary for the VIs of the recall and the VIs obtained from the conversation partners to *share the same scene*. Although the design of the Xapagy system in this respect is in flux, we assume that such a model would indeed be the default. The implication is that the agent would maintain only one instance of Hector and Achilles, involving both the VIs from its own recall and the VIs received from the conversation[17].

Figure 14 shows an example of how a leading statement can affect the recall. In the right side of the figure, we see the focus, with the VIs in white being previously recalled, while the VI in black being inserted externally, through a conversation. All the VIs will create support for various HLSs. An inserted statement does not necessarily change the flow of the recall: it might, for instance, support the same dominant story line, or the inserted HLS might support an HLS which has very low total support which will never be chosen for instantiation.

However, an external statement *can* change the flow of the recall if it supports an HLS which is already in close competition for the continuation, i.e. it is supported by other story lines in the agent's episodic memory. Thus, a successful leading statement (or question) is one which is based on existing tendencies of the agent (e.g. stereotypical continuations of certain stories).

In the figure, the inserted statement will increase the support of HLS B (which is also supported, albeit weakly, by the dominant story line). This additional support changes the flow of the recall from A → C → D to B → C → D.

*7) Case 7: Story line jumping:* Up to this point we assumed that the recall follows a single dominant story line, with potentially minor disturbances or deviations. Case 7 considers the situation where the agent switches the dominant story line, and starts to follow a new story line. For instance, an agent might follow the Iliad up to the point where Achilles kills Hector, then switch to the story of King Arthur and follow a story line where Achilles has a illicit love affair with the wife of Menelaus (that is, with Helen).

The switching of the story line maintains the scenes and instances. However, the shadows will be gradually taken over by the instances from the newly followed story line. The recall will still be about a warrior called Achilles - however, the shadow of the instance Achilles will be dominated by the in-memory instance of Lancelot.

Story line jumping is a natural extension of competitive recall. In Case 2 we considered situations where only a small number of additional HLSs have been received from the foreign story line before returning to the dominant story line.

In the current version of Xapagy, if the recall follows at least two-three HLSs from the foreign story line, the dominance will inevitably jump to the foreign story line, as the story consistency DA will reinforce the new story line. In our example, it is enough for the story line to shift from the Iliad to the Knights of the Round Table for several VIs to establish the Arthurian epic as the new dominant story line. This does not satisfactorily mimic the human story telling, which allows for much longer detours before returning to a main story line. In functionality which is currently under implementation, the summarization VIs would provide a greater stability of the story line over longer stretches of time. This would allow the recall to include relatively long elements of the foreign story, without abandoning the current story line. For example, with this functionality we would be able to insert a love affair between Achilles and Helen while still remaining in the general outlines of the Iliad.

*8) Case 8: Random wandering:* In our last considered case, the recall performs frequent story line jumps. In the extreme version of this case, the dominant story line can be different for every recalled VI (a situation illustrated in Figure 16). The recalled story will resemble no witnessed, heard or read story, although the individual VIs will still be instantiated based on an HLSs supported by VIs from the episodic memory. The agent can not recall a VI of "kissing", if it had not witnessed, heard or read about this action. However, HLS might have support coming from multiple story lines, with the various verbs averaging out in the instantiation template. Thus the recalled VI might not be an exact match of any previously encountered VI.

---

[17] The Xapagy architecture also allows a scenario in which the agent maintains two different instances in two different scenes. We can also envision linking these two instances with a special identity (conversational identity). This is consistent with the practice of the rest of the Xapagy system where we are using similar identity relations to maintain, for instance, somatic identity between instances undergoing change.

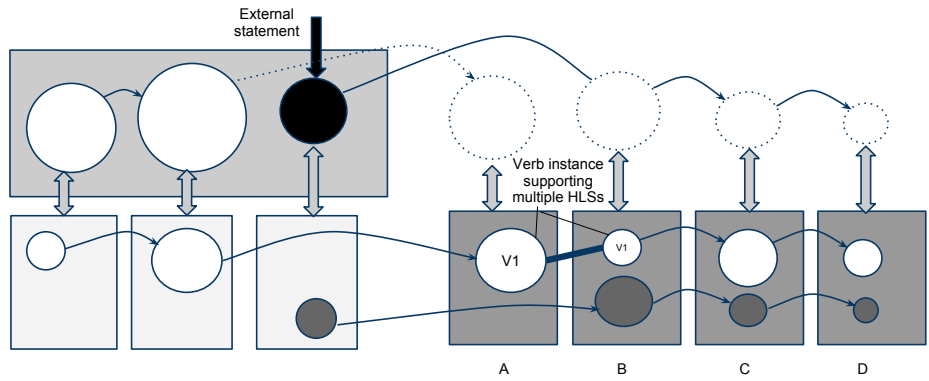

Fig. 14.    Case 6: Leading questions. Without the leading question, the recall will be A → C → D. The leading question however, brings into the shadow a foreign story line, which creates a different interpretation of HLS A, in HLS B (an interpretation which is still supported, albeit weakly, by the dominant story line). The recall will, thus, proceed along the lines of B → C → D.

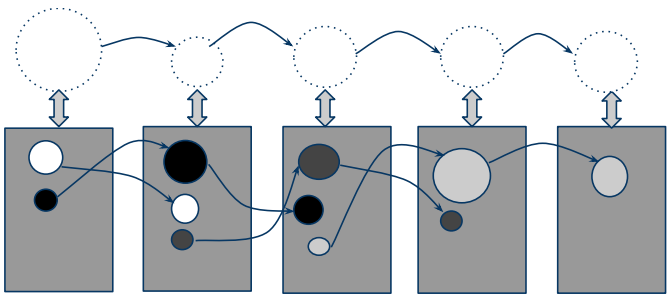

Fig. 16.    Random wandering

Case 8 can be used to mimic daydreaming and the creative process of original story writing. On the pathological side, it can mimic fantasy prone personality (FPP), Ganser's syndrome and other related disorders.

The Xapagy agent, in its default settings, will not perform such a random walk. However, its parameters can be tuned to random walk by:

- An inconsistent, possibly time-variable constraint filter.
- Lowering of the strength of the story consistency DA in the shadows.
- Introducing a counter-consistency DA, operating on the shadows and HLSs which reduce the support from story lines which have been recently followed.

### D. Divagations

*Hence the risk of what I would call Salgarism. When characters in Emilio Salgari's adventures escape through the forest, pursued by enemies, and stumble over a baobab root, the narrator suspends the action in order to give us a botany lesson on the baobab. Now this has become topos, charming, like the defects of those we have loved; but it should not be done.*

("The title and the meaning [of the Name of the Rose]" by Umberto Eco)

We will group under the term *divagations* all the recalled VIs which do not directly advance the story. The agent performs a divagation during recall if it does not instantiate a continuation HLS, instead it creates a VI through different means. The term divagation is not used in a pejorative sense: many divagations are necessary to maintain the flow of the recalled story, many of them are important in communication to allow the communication partner to create an accurate match of the recalled story[18]. We will consider three types of divagations:

- instance creation and initial characterization
- elaborating on instances
- justifying actions

In the following we shall describe these divagations in more detail.

*1) Instance creation and initial characterization:* A recalled story involves not only actions but also active and passive participants (warriors, swords, shields). We need to create instances in the current story for each of them, and for this we need explicit, `[create-instance]` type VIs.

The creation of the instances is mandatory: before the recalled story can first instantiate a VI using that instance, the instance *must* exist in the focus, and it needs to have a shadow which connects to the right HLS. Even if we recall the dominant story exactly, the instance creation still must be handled separately, because the `[create-instance]` VIs are *not* action VIs, they will not be instantiated through continuation HLSs[19].

The agent has considerable freedom with respect to *when* to create the instances, and what attributes will the instance initially receive. The agent can follow several strategies. The current implementation of the Xapagy system uses a *just-in-time* strategy: the continuation HLS is chosen and if one of the parts of its template maps to a non-existent instance, the instance is created before the HLS is instantiated.

An alternative approach (not currently implemented in the Xapagy system) would parse the current continuation chain, collect all the necessary instances, and instantiate them in a single burst of instantiation. As it happens, Homer uses an extreme variant of this approach: in the "catalogue of the ships" in the second book of the Iliad he lists

---

[18]This applies both to real time, interactive communication, as well as creating stories for later reading.

[19]Except in cases where the instance has been created through an action: i.e. it was made, born, appeared or it was created as a result of a change operation

all the 29 contingents of the Greeks led by 46 captains, introducing all the Greek characters of the epic at its very beginning.

A newly introduced instance must be immediately *characterized* with some attributes in order to be able to play its role in the recalled story. The memory instances shadowing the newly created instance have all the attributes they acquired up to the moment when they left the focus. It is as if the agent would know, at the first word of the retelling of the Iliad, all the attributes of, say, Ulysses (which, in fact, the human narrator *does* know). Only a subset of these attributes need to be assigned to the newly created instance immediately after creation. The Xapagy agent considers three criteria in the assignment of the initial attributes: (a) referential uniqueness, (b) conveying the "essence" and (c) shadow uniqueness.

**(a) The referential uniqueness criteria** requires that the newly introduced instance should have enough attributes to be uniquely referrable in the focus. The proper noun (if there is one) is a particularly good choice:

```
1   Scene / create-instance / "Ulysses".
```

but other choices are also good as long as they make the reference unique:

```
1   A crafty warrior / exists.
2   An ithacan warrior / exists.
```

The Xapagy agent must satisfy the referential uniqueness criteria if the story is *verbalized*, because in the Xapi form of the story, an instance without referential uniqueness can not be referred again. However, if the agent recalls a story internally, without verbalizing it, it is sufficient that the shadow of the instance to be sufficiently distinct, as that is the way in which the HLSs refer back to instances.

**(b) Conveying the "essence".** When introducing a new entity in natural language, we usually expect the narrator to make a characterization which, in some sense, conveys the "essence" of the entity. We can find examples of literary works, where this essence is not conveyed directly. Thomas Carlyle noted that his wife has read the poem Sordello by Robert Browning and had not been able to tell whether Sordello was a man, a city or a book. Nevertheless, most stories explicitly convey this meaning.

The current version of Xapagy can convey the "essence" by requiring that an instance is characterized by at least one concept near the "basic" level of the concept hierarchy (in the sense described in Rosch *et al.* [23]). In Xapagy, these are the concepts with an area close to 1.0.

For instance, saying that Ulysses is a man (area = 1.0) or that he is a warrior (area = 0.8) will satisfy the criteria. However, saying that he is alive (area = 5.0) is too wide, while simply assigning a narrow concept such as a proper name will also not be sufficient (e.g. [``Ulysses''] has an area 0.05).

It is interesting to observe that the referential uniqueness criteria favors concepts with a small area, while the essence criteria favors concepts of a specific area (around 1.0).

**(c) Shadow uniqueness criteria.**

For a story recall in progress, the shadow of the instances provide the link between the entities of the recalled story and the story lines which generated it. In principle, we would desire a unique mapping: a recalled instance must have shadows which contain exactly one instance for each of the competing story lines. This means, as well, that the shadow must be unique: it must be different from all the other current shadows.

From the point of view of characterization, the shadow uniqueness criteria becomes an issue because there is an interrelationship between the attributes of the instance and the shadow: adding an attribute to the instance will implicitly change the shadow as well (even if the attribute came from the shadow (for instance, through the matching the head DA).

*2) Elaborating on instances:* Elaborating on instances is the divagation through which an agent, instead of recalling action verbs which further the story, recalls is-a type VIs which add attributes to various instances in the focus.

As we have seen, a certain number of attributes must be assigned to the instance at its creation. Further attributes can be brought in later.

The current implementation works as follows. The system creates a number of *elaboration HLSs*, each containing an instance and an attribute, chosen from those attributes of the shadow components which do not already exist in the instance. The instantiation of such an HLS will generate a VI of type `I / is-a / C`.

The elaboration HLSs compete with the continuation HLSs and the other HLS types for the right to provide the next VI. The weight of the elaboration HLSs in this competition is affected the following factors:

- **Elaboration preference:** a continuous, slowly changing weighting factor which scales the weight of each elaboration HLS.
- **Gradient:** describes the difference between the attributes of the shadow and the instance. Each instantiated elaboration brings into the instance attributes present in the shadow, thus diminishing the gradient.
- **Saturation:** the weight of further elaborations is reduced as the number of attributes grow.
- **Momentum:** it is desirable to cluster elaborations in short bursts of elaborations for a unique instance. If a certain elaboration HLS had been chosen for instantiation, it creates a short term increase in the preference for elaboration HLSs for the same instance.
- **External attribute preferences:** an externally configured concept overlay provides a weighting factor of which concepts are preferred to be elaborated upon. This can not bring in attributes which do not exist in the shadow, but can suppress those which are preferred to be brought into the attributes of the current instance.

*3) Justifying actions:* Over time, the existence of verb instances in the memory of the agent generate *cloud relationships* among the concepts and the verbs. A cloud, for instance, describes the likely set of attributes of an instance which is the subject of a particular verb. For instance, warriors are expected to strike their opponents, while we would not expect a basket to do the same.

Justification is not mandatory. There is nothing to prevent an agent from instantiating a verb instance whose parts are not justified.

In some cases no justification is possible. An example is the wolf in the Little Red Riding Hood story: none of the known attributes of the wolf are compatible with talking - yet this does not prevent the narrative reasoning over the story.

### E. Story steering

*1) Initialization of the recall:* In the Xapagy system there are no specific markers of the beginning or the end of a story, nor a "name" of a story which allows it to separate from the remainder of the memory. If we want to trigger the recall of a specific story, we need to externally create an environment in which the recall can be started.

The current version of the Xapagy system uses *triggering by initial condition* - it requires the establishment of the initial condition of the story in order to start the recall.

```
1   Troy / is-only-scene.
2   A "Hector" / is-a / trojan warrior.
3   An "Achilles" / is-a / greek warrior.
4   "Hector" / hits / "Achilles".
5   Scene / recall narrate.
```

The first three lines establish a focus which contains Troy as the scene and establishes the presence of Hector and Achilles in the scene. The fourth line adds an action verb which will have continuations[20]

In line 5, the `[recall]` verb switches the Xapagy agent into recall mode. In this mode, the agent, instead of expecting VIs from outside observations, it will generate them internally. The recall mode terminates when the agent has decided that there is no sufficiently supported HLS to instantiate next. Whether this can ever happen, depends on the thresholds chosen and the parametrization of the HLS creation. For some parameter combinations, the agent will jump from one recalled story to another or to confabulate indefinitely.

The focus at the moment of starting the recall determines not only the first recalled VI, but the subsequent shape of the recall as well. If we want the agent to accurately recall a specific story, we need to initialize the agent with the most specific components, and parametrize the shadow maintenance processes to favor the strongest component. Otherwise, the shadows will contain elements from other stories, which then can shape the recalled narrative - or even cause it to divert to a different story line altogether.

*2) Story line sharpening:* is a technique through which during a competitive recall, we increase the lead of the dominant story line in the shadows and the HLSs. Story line sharpening is especially important when we need to choose between story lines whihc evolve in parallel but diverge in the middle of the story.

Let us consider the example of Hector again. As we have shown, we can consider:

- Hector the trojan warrior from the Iliad of Homer
- Hector the quantum recreation on the planet Mars from the novel Ilium by Dan Simmons.

Which of these stories we are actually narrating? We can actually narrate quite an extensive sequence of events without making a choice. However, if at some moment we make a decision and choose to elaborate on Hector with the statement:

```
1   "Hector" / is-a / quantum-recreation.
```

than this will affect the shadows: the interpretations based on Dan Simmons' novel will become stronger, and the future recalls will become more likely to follow the novel's story line (Hector forming an alliance with Achilles against the gods).

Note that it is possible for an agent to recall an attribute from another story line, yet it would not derail the story. For instance, the talking wolf in the Little Red Riding Hood will be shadowed by all the talking wolfs previously encountered, including the one from the "Nu zaec, pogodi!" Russian cartoon, where the wolf is a heavy smoker. It is possible to retell the story of Little Red Riding Hood with a heavy smoker wolf without slipping away completely from the story line. One reason for this is that in the cartoon, the wolf's nemesis is a bunny, which would not match well as a shadow for the red hooded girl.

## IX. RELATED WORK

The Xapagy system had been developed from scratch but not in a void. In the following we shall compare and contrast it against a small collection of representative systems. Most of these systems represent bodies of work which fit the Newell definition of cognitive architecture [19] as integrative systems which try to provide a model of a large subset of the human cognition even if they do not answer to all the twelve requirements of a human cognitive architecture as proposed in the "Newell test" (Anderson and Lebiere [3]). Some of them have not been explicitly positioned as cognitive architectures.

We shall look both at differences and similarities in the approach. The Xapagy system has taken a somewhat unusual approach to issues such as instance and concept as well as in focusing on episodic memory and shadowing as a single reasoning model. These choices have their root in the application, computational efficiency considerations - but also in the desire to try out roads less traveled. Despite of this, we found unexpected similarities and convergencies between Xapagy and approaches which have started out with radically different assumptions.

Xapagy is a new system, developed over the last three years and first described in this paper. When comparing it against other architectures we will also focus not only on what we learned from some of these systems but also what we might learn in the future.

*The ACT-R cognitive architecture*

ACT-R (Anderson and Lebiere [2], Anderson et al. [1]) is a cognitive architecture developed based on the theoretical models of John R. Anderson. It is the center of an active community of researchers with many variants adapted to perform specific tasks. Overall, the ACT-R community has been eager to integrate the system with external models. Earlier extensions, such as the ACT-R/PM (perceptual-motor) model have been integrated into the core of the ACT-R system. A recent integration effort created the SAL hybrid architecture through the integration of ACT-R and the Leabra neural model (Jilk et al. [9]).

The central idea of the ACT-R system is the use of the *production* as the model of the functioning of the human brain. A production requires some preconditions to be satisfied, and then it can "fire", performing a certain action. Unlike other models such as EPIC, ACT-R is based on a serial-bottleneck theory of cognition, where only one production can fire at a time. This is similar to the Xapagy system which also allows only one verb instance to be inserted into the focus at a time. Although the idea of a production is a computer science term, it has been found that it can be quite efficiently mapped to lower level terms, connectionist systems and even biological brain functions.

The flow of cognition in ACT-R happens through the system choosing productions based on its current *buffers* (largely equivalent to the Xapagy focus or the working memory in several other systems). The buffers can hold the current goal, its declarative memory, perceptual and motor modules, where it goes through the matching /selection / execution model.

ACT-R is a goal oriented system, it describes the active behavior of an agent. In general, ACT-R emphasizes the conceptual knowledge as encoded in the productions, and it does not focus on the episodic knowledge.

In contrast, for the Xapagy system, the episodic knowledge is the main focus. This is not really a choice, when we consider the fact that the objective of the Xapagy system is is the reasoning about stories.

*The SOAR cognitive architecture*

SOAR is a cognitive architecture with an active academic community and history of successful commercial application. In its initial version (Laird, Newell and Rosenbloom [12]) SOAR was primarily a symbolic production system. However, its recent development path increasingly integrates non-symbolic memory and processing components in the architecture (Laird [11]).

In its traditional version, SOAR has a long term memory (encoded as production rules) and a short term memory (encoded as a symbolic graph structure where objects are represented with properties and relations). The SOAR system performs cognition by matching and firing production rules. SOAR modulates the choice of the production to fire through operators. If the current preferences are insufficient for the choice of the production rule to fire, the SOAR system declares an *impasse*, automatically creating a new sub-state with the explicit goal to resolve it. This can be performed repeatedly and recursively, leading to an automatic meta-reasoning structure.

The recent version of the SOAR system extends the system in form of non-symbolic knowledge representations (and the implied memory and processing modules) as well as new learning models for knowledge not easily encodable in form of rules.

From the point of view of the relationship to the Xapagy system it is important that the new version of SOAR implements episodic memory in form of temporally ordered snapshots of working memory. Recorded episodes can be retrieved by the creation of a *cue* which is the partial specification of a working memory. The retrieved episodes are stored in a special buffer, the exact recall of a sequence of episodes is possible. The retrieval of episodes is also affected by a new module modeling the activation level of working memory.

Compared with the Xapagy architecture, the main difference is, again, the emphasis on the episodic memory placed by the Xapagy system. SOAR defines several memory models (procedural knowledge encoded as rules, semantic memory (facts about the world) with the episodic memory being a relatively recent addition. In contrast the episodic memory is essentially the only memory model in Xapagy[21].

The cue and working memory based episodic memory retrieval parallels the Xapagy's continuation HLS model. There is however a major difference in what the actual recall does. In SOAR it is possible to accurately recall a previous episode, with the complete snapshot of the working memory. In Xapagy this is impossible - it is not sure that even the general outline of the recalled story will be accurately recreated (see Section VIII-C).

### The ICARUS cognitive architecture

The ICARUS system (Langley and Choi [14]) is a cognitive architecture which focuses on cognition occurring in the physical context. Its cognitive behavior cycle is strongly tied to the perceptual input from the environment, with the mental structures grounded in the perception and the actions of the agent. ICARUS models *concepts* and *skills* as encoding different aspects of knowledge, memorized as different cognitive structures. It assumes that each element of the short term memory has a corresponding generalized structure in the hierarchically organized long term memory.

Comparing ICARUS and Xapagy is difficult, as the physical, grounded orientation of ICARUS and the story oriented application domain of Xapagy have very little overlap. Inevitably, however, Xapagy needs to extend towards application scenarios where the stories are recorded as an active participant in the events (rather than as a more or less passive observer of events or listener of narrated stories). Thus the ICARUS system can serve as inspiration for future Xapagy extensions for embodied action and goal directed reasoning.

### The Polyscheme cognitive architecture

The Polyscheme system (Cassimatis et al. [5]) is a cognitive architecture designed to integrate multiple different representations and reasoning algorithms. Polyscheme reasoning proceeds through sequential mental simulations where the different cognitive reasoning and planning models are encapsulated in *specialist* modules. The architecture allows even purely reactive components to react to simulated states of the world, thus integrating reasoning and planning with perception and action.

The integration of the different specialists is performed through a common communication language, used to translate information between specialists. Beyond this, however, each specialist can use its own internal data representations.

The Polyscheme system implements a rigorous and consistent way to annotate states as being in the past, hypothetical, distant or invisible.

There are apparently irreconcilable differences between the architectural approach of Polyscheme and Xapagy. Polyscheme is fundamentally built on the integration of multiple representations and reasoning models while Xapagy uses a single representation and a shadowing as a single reasoning model.

A closer look, however, also finds many parallels. The common language used between specialist modules can be seen as an analogue to the Xapagy verb instance representation.

Many of the specialist modules in Polyscheme would need to be also implemented outside the shadow/HLS reasoning model if Xapagy would be used in their application domain. For instance in the Polybot robot application one of the specialists identifies the location and category of the objects using color segmentation. This obviously cannot be implemented through shadowing - should the Xapagy system ever be controlling a robot, the solution would roughly be the same: a specialized visual module which communicates with the rest of the system by inserting instances and verb instances into the focus.

Finally, the shadow/HLS mechanism in Xapagy can be considered a type of simulation of recent events based on past knowledge, which is a close parallel to what the Polyscheme simulation model does.

### The EPIC cognitive architecture

The EPIC cognitive architecture (Kieras and Meyer [10]) has been implemented with the explicit goal to model the human multimodal and multiple task performance. The EPIC architecture's goal is to understand and model human behavior (preferably, through a predictive model) rather than to create an artificially intelligent agent.

As ACT-R and SOAR, EPIC is a production based system. One of the influential differences of the EPIC system is that it does not assume a serialized bottleneck on the firing of the productions. However, the number of productions which can be actually be instantiated simultaneously is limited by resource limitations, for instance the various components of the working memory, the processing speed, and the physical limitations of sensors and actuators (eyes can focus on one point at a time, hands can grab a single object, and so on).

What the Xapagy system can learn from EPIC is the lesson which many other architectures have also drawn - that is, that the perceived serialization of some aspects of human behavior can occur naturally - they need not be enforced through artificial constraints.

### The CLARION cognitive architecture

The stated goal of the CLARION architecture (Sun [28]) is to capture cognitive modeling aspects not adequately addressed by other cognitive architectures: the implicit-explicit interaction, the cognitive-metacognitive interaction and the cognitive-motivational interaction.

---

[21]Xapagy also defines conceptual and dictionary knowledge. Conceptual knowledge is implicitly present in SOAR (and basically all the other cognitive architectures, even if they do not appear a separate module). Dictionary knowledge is also necessarily present in any system which uses some kind of external language.

Overall, CLARION is an integrative model with separate functional subsystems (such as the action centered, metacognitive and motivational subsystem). Each subsystem has a has a dual representational structure, separately for implicit and explicit representations. This separation of the high level (symbolic, explicit) and low level (non-symbolic, implicit) representations is consistently carried through throughout the architecture. These two representation models are independently operating and compete for the right to perform an action through well defined interaction models. For instance, the system contains models for learning symbolic rules from subsymbolic information. The CLARION system also contains built-in motivational and metacognitive constructs.

On a superficial look, there appears to be very little similarity between CLARION and Xapagy. CLARION's characteristic is that its architecture explicitly models several dichotomies of human cognition (explicit vs. implicit, action vs. deliberation, symbolic vs. subsymbolic). If one wishes to identify where a certain human reasoning ability is modeled, a single look at the CLARION's architectural diagram can point to the component responsible. In contrast, Xapagy uses a single reasoning model, shadowing, and whatever dichotomies must appear in the external behavior, it must emerge from a unique internal model.

Yet, despite the very different architecture, a careful comparison study can find that parallels between the turn architectures can be drawn. For instance, CLARION uses a mix of similarity based and rule-based reasonings - these components can be mapped to the various shadow maintenance activities which deploy a mix of structural and similarity based matching techniques.

### The KARMA story understanding system and related work

The KARMA system for story understanding (Narayanan [17]) had demonstrated the understanding of simple stories and narrative fragments in the domain of international politics and economics as described in journalism.

One feature of the system is the ability to represent, reason about, and understand metaphors which present abstract concepts (such as those in international politics and economy), in terms of the physical motion, forces, terrain features. It uses the Event Structure Metaphor that projects inferences from physical motion and manipulation to abstract actions, goals and policies.

From the point of view of practical implementation, the KARMA system uses a model called *x-schemas* which are parametrized routines, with internal state which execute when invoked. They are implemented as an extension to Petri-nets (to allow run-time binding, hierarchical action sets and stochastic transitions).

The KARMA system is strongly focused on the study of *aspect* the study of linguistic devices which profile and focus on the internal temporal character of the event.

A recent paper (Lakoff and Narayanan [13]) described a roadmap towards a system which understand human narratives by exploiting the cognitive structures of human motivations, goals, actions, events and outcomes. The authors outline a number of directions towards which future narrative systems need to work, and order them in *dimensions of narrative structure*: the moral system of the narrative (stories as guides to living), folk theories of how people and things work, overall plot structures (these are in many way resemble summarizations in Xapagy), plot schemas, motif structures, and narrative variations.

### The DAYDREAMER system

The DAYDREAMER system (Mueller and Dyer [15]) was a system which generated stories, based on, and initiated from actual experiences. The system used in the guidance of these story creations four goals which were identified as frequently appearing in daydreaming: rationalization, revenge, failure/success reversal (exploration of alternatives), and preparation (generation of hypothetical future scenarios).

DAYDREAMER had been strongly tied to models of emotion: defined notions of success of a goal (which produces a positive emotion) and failure (producing negative emotion). It proposed that the basic mechanism for scenario generation is plannning.

The DAYDREAMER system was taking a similar approach to Xapagy in the fact that both the actual experiences and the daydreamed experiences are equally available to recall.

DAYDREAMER modeled a dynamic episodic memory (Tulving [29], Schank [24]), constantly modified during daydreaming. In Xapagy, the episodic memory is not, in fact, modified, yet the practical impact is similar, as the way in which the recall happens through self-shadowing, it creates a behavior compatible with a dynamic memory for the external observer.

### Recent work on narrative schemas by Chalmers and Jurafski

An body of work, with a high relevance to the Xapagy system has been recently reported by Chambers and Jurafsky [6], [7]. The work centers on the learning of narrative schemas, which they define as coherent sequences or sets of events with participants who are defined as semantic roles, which can be fitted by appropriate persons.

This work can bee seen as a restart of the semantic NLP tradition of the 1970 and 80s, which used representations such as scripts (Shank and Abelson [25]). This research direction had become marginalized in the 1990's as the knowledge engineering requirements appeared to be overwhelming. The work of Chalmers and Jurafski promises to reboot this research direction as the narrative schemas can be inferred or learned in an unsupervised way from raw input.

In [6], [7] successfully show scema and role inference, starting from preprocessed natural language. The evaluate the system using the *narrative cloze* test, an variation of the classical word-based cloze test, where an event is removed from a known narrative chain, and the system needs to infer the missing event.

The work is very close to the Xapagy system in its external objectives, as well as in some of the details of the processes (e.g. some of the shadow maintenance DAs are close relatives of the algorithms used by the authors). It is true, Chambers and Jurafsky start from English text, not pidgin, but they are using independent preprocessing algorithms to resolve the references (albeit with some level of uncertainty). Practically, the input of the algorithms is closely related to the verb instance form of the Xapagy input.

There is, however, a difference between the two approaches, which, depending on how we look at it, is either a major philosophical difference or an implementation detail. The assumption of Chambers and Jurafsky is that *scripts physically exist*, that the agents have an explicit data structure, which is, let us say, the restaurant schema, which is the subject of knowledge engineering or learning.

In contrast, in Xapagy, such an internal scheme does not exist as an explicit data structure. The reasoning model of Xapagy (shadows and headless shadows) do not create permanent intermediary structures: reasoning always starts with the concrete, originally memorized story lines.

Looking from another perspective, however, the differences are not as great. For instance, a Xapagy agent, asked to confabulate a restaurant scene based on generic participants, will recall the same sequence as the one which one would obtained from a narrative schema initialized with the same participants in the specific roles. So, for an external observer, which sees the system's behavior but not the internal structure, these systems would be hard to distinguish (at least as long as we consider typical scenarios - for atypical scenarios, exceptions and borderline cases there is a larger freedom of movement for both systems).

## X. Conclusion

The Xapagy system is an active software project, in development since 2008. It has been designed to perform narrative reasoning, that is, to mimic the human behavior when reasoning about stories. The objective of this paper was to convey the architecture of the system, with an emphasis on the design decisions which differentiate it from its peers from the artificial intelligence literature.

In order to fit the description of the architecture in a single paper, we had to omit certain details. For instance, the spike activities and diffusion activities which maintain the shadows and headless shadows have been presented in an informal manner. The various narrative reasoning modalities outlines in Section VIII would deserve separate and extensive treatment, which will be provided in future papers.

## Appendix A
## Abbreviations, notations and typographical conventions

This paper uses a relatively extensive list of mathematical notations to describe the interaction between the different components of the system.

We are also using specific typographical conventions to show the various levels of input, intermediary values and output of the system. These values will be enumerated here.

### A. Acronyms

The following acronyms are used in the paper:

VI   verb instance
CO   concept overlay
VCO   verb concept overlay
SA   spike activity
DA   diffusion activity
HLS   headless shadow

### B. Notations for basic components

$V$ a verb instance
$I$ an instance
$c_i$ a concept (an attribute or adjective)
$v_i$ a verb (or adverb)
$-c_i, -v_i$ the negation of a concept or verb
$C$ an overlay of concepts
$CV$ an overlay of verbs
$area(c)$ area of a concept (similar for a verb)
$overlap(c_i, c_j)$ the overlap between concepts $c_i$ and $c_j$
$impact(c_i, c_j)$ the impact of concept $c_i$ over concept $c_j$
$een(C, c)$ explicit energy of concept $c$ in overlay $C$ (similar for the verbs)
$en(C, c)$ implicit energy of concept $c$ in a overlay $C$ (similar for the verbs)
$act(C, c)$ activation of concept $c$ in overlay $C$ (similar for the verbs)

### C. Notations for reference resolution

$match(C_1, C_2)$ the match between two concept overlays
$match(V_1, V_2)$ the match between two verb overlays

### D. Notations for verb instances

$vitype(V)$ the type of the verb instance $V$ (can be S-V-O, S-V, S-ADJ, S-ADV, IND)
$comp(V)$ the list of components of the verb instance $V$
$Verb(V)$ the verb overlay which defines the verb instance $V$
$SubI(V)$ the instance which serves as the subject of the verb instance $V$ (if type S-V-O, S-V, S-ADJ, IND)
$ObjI(V)$ the instance which serves as the object of the verb instance $V$ (if type S-V-O)
$ObjCO(V)$ the concept overlay which serves as the adjective of the verb instance $V$ (if type S-ADJ)
$SubVI(V)$ the verb instance which serves as the subject of the verb instance $V$ (if type S-ADV)
$ObjVO(V)$ the verb overlay which serves as the object adverb of the verb instance $V$ (if type S-ADV)
$pred(V_1, V_2)$ the strength of the predecessor relationship from verb instance $V1$ to $V2$
$succ(V_1, V_2)$ the strength of the successor relationship from verb instance $V1$ to $V2$

### E. Notations for focus and shadow

$sh(I_F, I_S)$ the participation of the instance $I_S$ in the shadow of the focus instance $I_F$
$sh(V_F, V_S)$ the participation of the verb instance $V_S$ in the shadow of the focus verb instance $V_F$
$M_I(I_F, I_S)$ the intrinsic matching between instance $I_F$ in the focus and instance $I_S$ in the shadow
$M_C(I_F, I_S)$ the contextual matching between instance $I_F$ in the focus and instance $I_S$ in the shadow
$M(I_F, I_S)$ the overall matching between instance $I_F$ in the focus and instance $I_S$ in the shadow
$M_{SC}(V_F, V_S)$ the semi-contextual matching between verb instance $V_F$ in the focus and verb instance $V_S$ in the shadow
$M(V_F, V_S)$ the overall matching between between verb instance $V_F$ in the focus and verb instance $V_S$ in the shadow

## References

[1] J. Anderson, D. Bothell, M. Byrne, S. Douglass, C. Lebiere, and Y. Qin. An integrated theory of the mind. *Psychological review*, 111(4):1036, 2004.

[2] J. Anderson and C. Lebiere. *The atomic components of thought*. Lawrence Erlbaum, 1998.

[3] J. Anderson and C. Lebiere. The Newell test for a theory of cognition. *Behavioral and Brain Sciences*, 26(05):587–601, 2003.

[4] S. Brubacher, U. Glisic, K. Roberts, and M. Powell. Children's ability to recall unique aspects of one occurrence of a repeated event. *Applied Cognitive Psychology*.

[5] N. Cassimatis, J. Trafton, M. Bugajska, and A. Schultz. Integrating cognition, perception and action through mental simulation in robots. *Robotics and Autonomous Systems*, 49(1-2):13–23, 2004.

[6] N. Chambers and D. Jurafsky. Unsupervised learning of narrative event chains. *Proceedings of ACL-08: HLT*, pages 789–797, 2008.

[7] N. Chambers and D. Jurafsky. Unsupervised learning of narrative schemas and their participants. In *Proceedings of the Joint Conference of the 47th Annual Meeting of the ACL and the 4th International Joint Conference on Natural Language Processing of the AFNLP: Volume 2-Volume 2*, pages 602–610. Association for Computational Linguistics, 2009.

[8] J. Hudson and K. Nelson. Repeated encounters of a similar kind: Effects of familiarity on children's autobiographic memory. *Cognitive Development*, 1(3):253–271, 1986.

[9] D. Jilk, C. Lebiere, R. O'Reilly, and J. Anderson. SAL: An explicitly pluralistic cognitive architecture. *Journal of Experimental & Theoretical Artificial Intelligence*, 20(3):197–218, 2008.

[10] D. E. Kieras and D. E. Meyer. An overview of the EPIC architecture for cognition and performance with application to human-computer interaction. *Human–Computer Interaction*, 12(4):391–438, 1997.

[11] J. Laird. Extending the Soar cognitive architecture. In *Proceedings of the 2008 conference on Artificial General Intelligence*, pages 224–235. IOS Press, 2008.

[12] J. Laird, A. Newell, and P. Rosenbloom. Soar: An architecture for general intelligence. *Artificial Intelligence*, 33(1):1–64, 1987.

[13] G. Lakoff and S. Narayanan. Toward a Computational Model of Narrative. In *2010 AAAI Fall Symposium Series*, 2010.

[14] P. Langley and D. Choi. A unified cognitive architecture for physical agents. In *Proceedings of the National Conference on Artificial Intelligence*, volume 21, page 1469. Menlo Park, CA; Cambridge, MA; London; AAAI Press; MIT Press; 1999, 2006.

[15] E. Mueller and M. Dyer. Towards a computational theory of human daydreaming. In *Proceedings of the Seventh Annual Conference of the Cognitive Science Society*, pages 120–129. Citeseer, 1985.

[16] T. Nagel. What is it like to be a bat? *The Philosophical Review*, 83(4):435–450, 1974.

[17] S. Narayanan. *KARMA: Knowledge-based active representations for metaphor and aspect*. PhD thesis, University of California, Berkeley, 1997.

[18] U. Neisser. John Dean's memory: A case study. *Cognition*, 9(1):1–22, 1981.

[19] A. Newell. *Unified theories of cognition*. Harvard Univ Pr, 1994.

[20] H. Noonan. Stanford encyclopedia of philosophy: Identity. URL http://plato.stanford.edu/entries/identity/, 2010.

[21] E. T. Olson. Stanford encyclopedia of philosophy: Personal identity. URL http://plato.stanford.edu/entries/identity/, 2010.

[22] S. Pinker. *Learnability and cognition: the acquisition of argument structure*. Cambridge, Mass.: MIT Press, 1989.

[23] E. Rosch, C. Mervis, W. Gray, D. Johnson, and P. Boyes-Braem. Basic objects in natural categories. *Cognitive psychology*, 8(3):382–439, 1976.

[24] R. Schank. *Dynamic memory: A theory of reminding and learning in computers and people*. Cambridge University Press, Cambridge [Cambridgeshire]; New York, 1982.

[25] R. Schank and R. Abelson. *Scripts, plans, goals and understanding: An inquiry into human knowledge structures*, volume 2. Lawrence Erlbaum Associates Hillsdale, NJ, 1977.

[26] M. Sebba. *Contact languages: Pidgins and creoles*. Palgrave Macmillan, 1997.

[27] A. Seikel, D. King, and D. Drumright. *Anatomy and physiology for speech, language, and hearing*. Delmar Cengage Learning, 2009.

[28] R. Sun. The CLARION cognitive architecture: Extending cognitive modeling to social simulation. *Cognition and multi-agent interaction: From cognitive modeling to social simulation*, pages 79–99, 2006.

[29] E. Tulving. Episodic and semantic memory. *Organization of memory*, pages 381–402, 1972.

[30] S. Zilberstein. Using anytime algorithms in intelligent systems. *AI magazine*, 17(3):73, 1996.